\pdfoutput=1

\documentclass[11pt]{article}
\usepackage[table,xcdraw]{xcolor}
\usepackage{hyperref}
\usepackage[final]{acl}
\usepackage{float}
\usepackage{hyperref}
\usepackage[normalem]{ulem}
\useunder{\uline}{\ul}{}
\usepackage{times}
\usepackage{times}
\usepackage{latexsym}
\usepackage{comment}
\usepackage{subfigure, caption, subcaption}
\usepackage{graphicx}
\usepackage{multirow}
\usepackage{xcolor}
\usepackage{xspace} 
\usepackage{amsmath}
\def\eg{e.g.,~} 
\def\ie{i.e.,~} 
\DeclareMathOperator*{\E}{\mathbb{E}}

\newcommand{\move}[1]{\textcolor{black}{#1}}

\usepackage[T1]{fontenc}

\usepackage[utf8]{inputenc}

\usepackage{microtype}

\usepackage{inconsolata}

\usepackage{graphicx}
\usepackage[bottom,flushmargin]{footmisc}
\usepackage{hyperref}
\usepackage{arydshln}

\usepackage{booktabs}
\usepackage{multirow}
\usepackage{colortbl}
\usepackage{amsmath}
\usepackage{xcolor}
\colorlet{tablerowcolFor}{gray!10} 
\newcommand{\formulation}[1]{$P_{#1}$\xspace}
\newcommand{\formulationset}{$P$\xspace}
\newcommand{\promptiog}[4]{$p_{#1}^{#2#3#4}$\xspace}
\newcommand{\ZS}{$\sigma_{zs}$\xspace}
\newcommand{\PO}{$\sigma_{0}$\xspace}
\newcommand{\Pf}[1]{$\sigma_{#1}$\xspace}
\newcommand{\Pall}{$\sigma_{0:3}$\xspace}
\newcommand{\contrastive}{$\sigma^{{0:3}}_{0}$\xspace}
\newcommand{\contrastiveO}{\contrastive}
\newcommand{\POO}{\PO}
\newcommand{\scenario}{$\sigma^{P_{c}}_{P_{ppo}}$}
\newcommand{\ppo}{{\sc ppo}\xspace}

\author{
 \textbf{Mohamed Salim Aissi\textsuperscript{1}},
  \textbf{Clement Romac\textsuperscript{2,3}},
  \textbf{Thomas Carta\textsuperscript{3}},
  \textbf{Sylvain Lamprier\textsuperscript{4}},
  \\
  \textbf{Pierre-Yves Oudeyer\textsuperscript{3}},
  \textbf{Olivier Sigaud\textsuperscript{1}},
  \textbf{Laure Soulier \textsuperscript{1}},
  \textbf{Nicolas Thome \textsuperscript{1,5}}
\\
  \textsuperscript{1}Sorbonne Université, CNRS, ISIR, F-75005 Paris, France
    \textsuperscript{2}Hugging Face
  \\
\textsuperscript{3}Inria (Flowers), University of Bordeaux, France
  \textsuperscript{4}Univ Angers, LERIA, Angers, France \\
  \textsuperscript{5}Institut universitaire de France (IUF)
  \\ \{salim.aissi, laure.soulier, olivier.sigaud, nicolas.thome\}@isir.upmc.fr
\\   \{clement.romac, thomas.carta, pierre-yves.oudeyer\}@inria.fr
  \\ \{sylvain.lamprier\}@univ-angers.fr
}

\usepackage[utf8]{inputenc} 
\usepackage[T1]{fontenc}    
\usepackage{hyperref}       
\usepackage{url}            
\usepackage{booktabs}       
\usepackage{amsfonts}       
\usepackage{nicefrac}       
\usepackage{microtype}      
\usepackage{xcolor}         

\title{Reinforcement Learning for Aligning Large Language Models Agents with Interactive Environments: Quantifying and Mitigating Prompt Overfitting}

\begin{document}

\maketitle
\begin{abstract}
Reinforcement learning (RL) is a promising approach for aligning large language models (LLMs) knowledge with sequential decision-making tasks. However, few studies have thoroughly investigated the impact on LLM agents capabilities of fine-tuning them with RL in a specific environment. In this paper, we propose a novel framework to analyze the sensitivity of LLMs to prompt formulations following RL training in a textual environment.
Our findings reveal that the performance of LLMs degrades when faced with prompt formulations different from those used during the RL training phase. Besides, we analyze the source of this sensitivity by examining the model's internal representations and salient tokens. Finally, we propose to use a contrastive loss to mitigate this sensitivity and improve the robustness and generalization capabilities of LLMs.

\end{abstract}

\section{Introduction}
\begin{figure}[t]
    \centering
    \includegraphics[width=1\columnwidth]{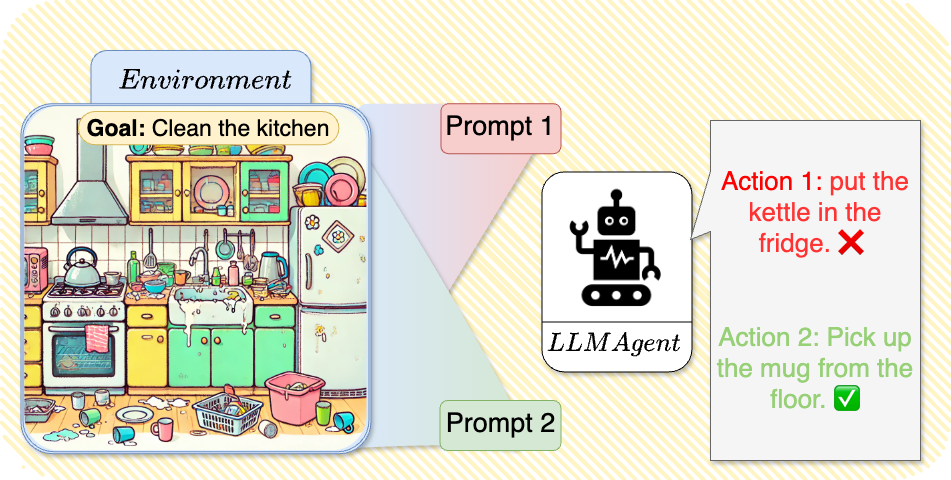 }
    \caption{\textbf{Sequential decision making with LLMs.} An LLM agent interacts with an environment by means of a prompt, \ie a textual representation of the scene contextualizing its goal, state description, and possible actions. Depending on the prompt, the LLM agent might choose different actions. In this paper, we show that LLMs fine-tuned with reinforcement learning tend to overfit to the specific prompts they have been trained on, and propose solutions to mitigate this effect.}
    \label{fig: LLM env}
\end{figure}

LLMs have demonstrated emergent abilities in tasks like summarization, translation or chain-of-thought reasoning~\citep{singhal2023towards, yao2024tree, wei2022emergent}. Their versatility suggests they possess some common sense knowledge and reasoning capabilities~\citep{li2021systematic, huang2022towards}, making them potential agents for tasks such as sequential decision-making~\citep{zeng2023large, zhao2024large, yao2022react}. In this context, LLMs predict actions based on a prompt, \ie a textual description of the  agent's goal and of the environment, \eg the agent's state and its possible actions (see \figurename~\ref{fig: LLM env}).

\noindent However, LLMs often face grounding issues, i.e. they suggest actions that may be unsuitable in the current situation. This is mainly due to the inadequacy of the common knowledge embedded in the LLM to the components or dynamics of the environment at hand~\citep{ahn2022can, huang2023grounded}. 
Solutions to this challenge include combining language with other modalities into multimodal models~\citep{jiang2022vima, driess2023palm, li2023mastering}, using dedicated grounding modules~\citep{huang2023grounded, yang2023llm}, or employing in-context learning (ICL) to correct prompts after spotting mistakes~\citep{ shinn2023reflexion, yao2023retroformer}. To explicitly integrate the specificities of the environment into the LLM itself, an emerging strategy involves using reinforcement learning (RL) to fine-tune LLMs and integrate knowledge from executing actions~\cite{carta2023grounding, tan2024true}. While those methods improve agent performance, it remains unclear how RL impacts the inherent knowledge of the LLM and whether the model gains generalization capabilities.

\noindent
In this paper\footnote{\footnotesize
Code can be found \href{https://github.com/sal1717lim/Reinforcement-Learning-for-Aligning-Large-Language-Models-Agents-with-Interactive-Environments}{here}}, we study the sensitivity of LLMs to prompt formulations and its impact on knowledge acquisition. We define a set of different prompt formulations to evaluate LLM performance when prompts are changed, and also analyze how the LLM represents these prompts.
\noindent Our findings reveal that the performance of LLMs is highly sensitive to prompt variations, suggesting that fine-tuning only induces superficial updates and fails to improve the acquisition of new knowledge about the environment.
~By analogy with observational overfitting in RL~\citep{song2019observational}, we refer to this phenomenon as {\em prompt overfitting}.
With this in mind, our paper proposes two main contributions:

\noindent$\bullet$ We design experiments to measure prompt overfitting issues of LLMs in interactive environments. 
The study reveals that fine-tuned LLMs heavily depend on the training prompts, exhibiting a significant drop in "zero-shot" performance when using new prompt formulations. To further analyze this behavior, we thoroughly analyze latent representations and salient tokens in LLMs, both showing a strong bias towards the prompt formulation. 

\noindent$\bullet$ We propose a solution for mitigating prompt overfitting with a contrastive regularization loss that makes the latent representation of the LLM invariant to prompt formulations. This solution significantly improves the zero-shot performance and the robustness to prompt variations, as well as the acquisition of new knowledge about the environment.

\noindent Altogether, our work contributes to a better understanding of the obstacles one must face when leveraging RL to improve the abilities of LLM agents to act appropriately in interactive environments.
\section{Related work}
\paragraph{Grounding LLMs for sequential decision-making.}
When solving sequential decision-making tasks, an LLM acting as an agent must leverage its common sense knowledge and adapt to the environment it interacts with. Several approaches have been explored to align the LLM with its environment. 
A first approach, called SayCan~\citep{ahn2022can} filters out inefficient actions, depending on affordances captured from interactions. The resulting agents can solve sequential decision-making problems, but the involved LLM does not benefit from interaction feedback. 
Alternatively, research has focused on functionally grounding LLMs in interactive environments using online RL to align text processing with external dynamics~\citep{carta2023grounding, tan2024true, wen2024entropy, zhou2024archer, abdulhai2023lmrl}. For instance, the pioneering GLAM method~\citep{carta2023grounding} enables LLM agents to learn policies through environmental interactions. 
Similarly, \citet{szot2024llarp} apply RL with Visual Language Models (VLMs) to solve embodied tasks but do not directly ground the VLM. Instead, they train randomly initialized neural networks on top of the VLM.
Parallel to these, several works~\citep{wang2023describe, wang2023voyager, wang2023jarvis} leverage prompting methods to inform the LLM about the consequences of its actions in the environment or even mix RL and prompting~\citep{yan2023ask}. Finally, in \citet{xiang2023language}, an LLM agent collects embodied experiments in an environment to enhance its modeling abilities such as counting and tracking objects, proposing plans, etc.
However, most existing works only employ a single prompt formulation, assuming that the LLM will adapt to the given format. In contrast, our study evaluates LLM performance across multiple prompts, aiming to understand how the model handles variations in formulation.

\paragraph{LLMs' Prompt Sensitivity.}
LLMs have shown remarkable capabilities to generate text and solve tasks in zero-shot and few-shot scenarios. To enhance their performance and consistency, various prompting methods have been developed \cite{liu2021pretrain}. 
In addition, multiple studies have highlighted the sensitivity of LLMs to minor perturbations in the prompt, leading to substantially different outputs~\citep{zhao2021calibrate, sclar2023quantifying, salinas2024butterfly}.
This sensitivity impairs the reliability and robustness of these models.  Indeed, certain input-agnostic sequences could trigger specific outputs, further illustrating the brittleness of LLMs to prompt modifications~\citep{wallace-etal-2019-universal}.
The sensitivity of LLMs persists regardless of model size, the number of examples, or the type of instruction provided~\citep{sclar2023quantifying, zhao2021calibrate, Loya_2023}. These studies also reveal poor performance consistency across models using the same prompt. To improve the LLM outputs, a recalibration can estimate and adjust for the model's biases with additional parameters, which mitigates the effects of prompt sensitivity~\citep{zhao2021calibrate}. Our work extends this research by evaluating the performance of LLM agents across various prompt formulations and by mitigating prompt overfitting to preserve semantic consistency in interactive environments. 

\section{Problem Statement and Methods} 
\subsection{Problem statement}
To investigate the impact of RL on the knowledge of LLM agents in interactive environments, we operate in a textual RL setting. Given a vocabulary of tokens~$V$, our experimental  framework  can be conceptualized as a partially observable Markov decision process~$M=(S,A,T,R,G,O,\gamma)$ with~$S$ the state space,~$A \subset V^{N}~$ the action space,~$G \subset V^{N}$ the goal space,~$T:S\times A \mapsto S$ the transition function,~$R:S\times G \mapsto S$ the goal-conditioned reward function,~$Obs:S\mapsto V^{N}\equiv O$ the observation function that maps a state~$s$ to a textual description and~$\gamma$ the discount factor. For a trajectory also called  rollout in RL ~$\tau=(s_1,a_1,\ldots,s_H,a_H)$ of length~$H$, we note~$R_g(\tau)=\sum_{t=1}^H \gamma^t R(o,g)$ its cumulative discounted reward given a goal~$g \in G$.  The optimal policy is~$\pi^*=\arg\max_\pi \mathbb{E}_{g \in G, \tau \sim \pi(\tau|g)}[R_g(\tau)]$, with~$\pi(\tau|g)$ the probability of~$\tau$ following~$\pi$.   
\begin{figure}[t]
    \centering
    \vspace{-0.5cm}
    \includegraphics[width=0.95\linewidth]{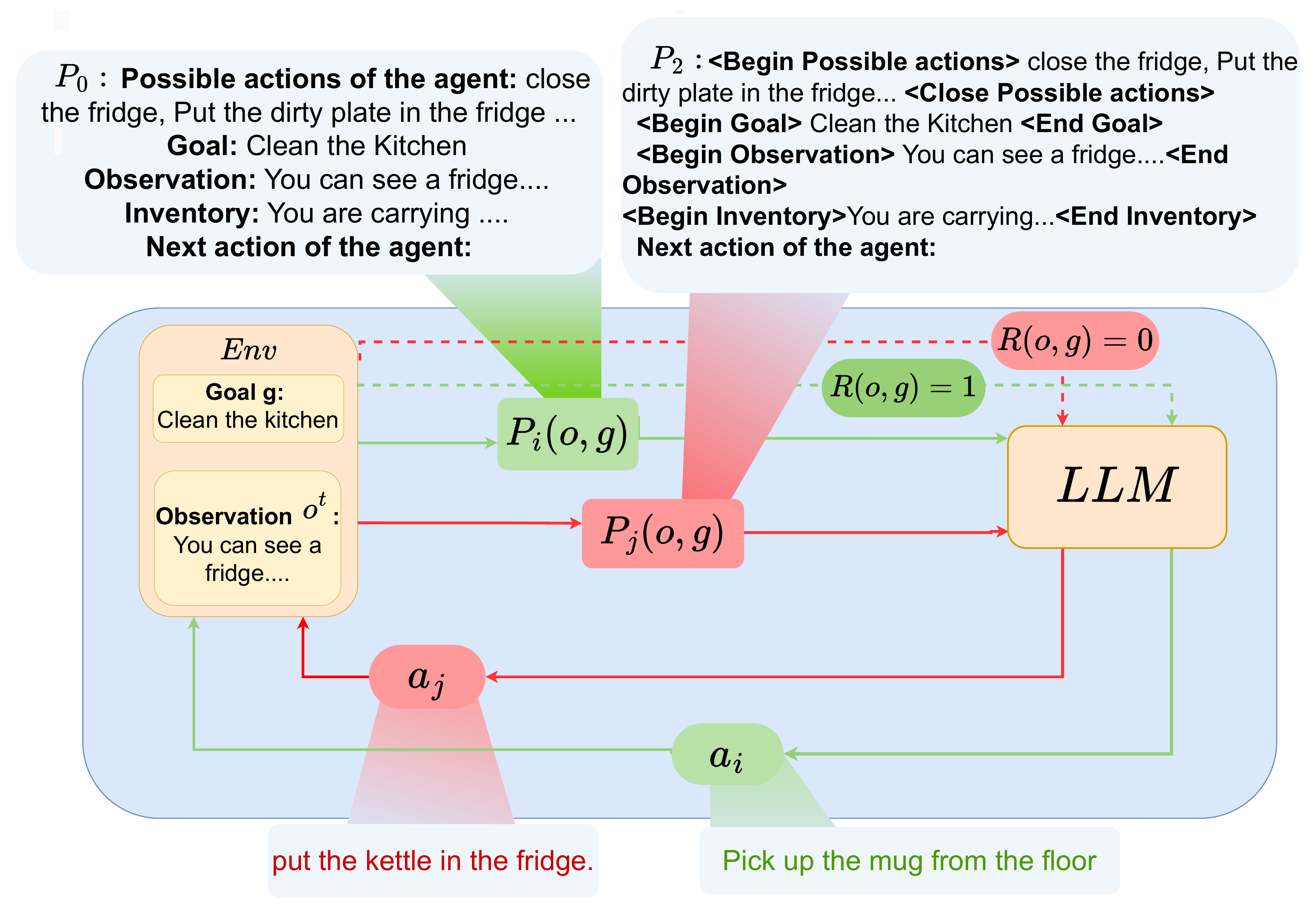}
    \caption{\textbf{The fine-tuning framework}: we use an LLM as an agent policy in a textual environment $Env$. The $Env$ provides a fixed goal description \(g\) for the current episode, a description of the agent's observation \(o\), and a scalar reward \(R(o,g)\) for the current step. The goal and observation are formatted using a prompt formulation \formulation{i}. Our experiments reveal that an LLM fine-tuned on \formulation{i} succeeds with prompts formatted with \formulation{i} but fails with prompts formatted with \formulation{j} $\neq$  \formulation{i}.
 } 
    \label{fig:main GLAM}
\end{figure}
\noindent On top of the above framework, we study the impact of prompt formulations \formulationset~$=\{P_i\}_{i \in \{1, \cdots, n\}}$, where~$P_i: O \times G \mapsto \cal P$$ \subset V^{N}$ formats text entries from observations and goals as prompt inputs.
We assume that all formulations from \formulationset~preserve information, i.e. any optimal policy~$\pi_i^*$ using the prompt formulation~$P_i$ from~\formulationset can obtain the same amount of rewards as an optimal policy~$\pi^*$ acting on original observations and goals:
~$\forall g \in G, \forall i \in \{1, \cdots, n\}, \mathbb{E}_{\tau \sim \pi_i^*(\tau|g)}[R_g(\tau)] = \mathbb{E}_{\tau \sim \pi^*(\tau|g)}[R_g(\tau)]$.

Based on this, we analyze the sensitivity of the policy $\pi$ to variations in prompt formulation compared to the one used during training. We define prompt overfitting as a scenario where, for a given policy $\pi$ trained using a specific formulation \formulation{i}, there exists a prompt formulation \formulation{j}$\in P$ such that the expected reward for $\pi$ on \formulation{i} is significantly higher than the expected reward for $\pi$ on \formulation{j}.

\subsection{Fine-tuning LLM agents with RL}
First, we define a policy $\pi$, based on an LLM for solving %
tasks in~$M$, given any prompt formulation \formulation{i}~$\in$ 
\formulationset. For any goal $g \in { G}$ and observation $o\in O$, we note~$\mathbb{P}_{LLM}(w_k|$\promptiog{i}{o}{,}{g}$,w_{<k })$ the probability of token~$w_k$
for prompt~\promptiog{i}{o}{,}{g} generated by the prompt formulation \formulation{i} and the decoding history~$w_{<k }$. We also define~$\mathbb{P}_{LLM}(a|$\promptiog{i}{o}{,}{g}$)=\Pi_{k=0}^{|a|} \mathbb{P}_{LLM}(w_k|$\promptiog{i}{o}{,}{g}$,w_{<k })$ the decoding probability of the sequence corresponding to action~$a~\in~A$ given prompt~\promptiog{i}{o}{,}{g}. Following~\citep{tan2024true}, we use a normalized decoding probability~$\mathbb{P}^{norm}_{LLM}(a|$\promptiog{i}{o}{,}{g}$)=\mathbb{P}_{LLM}(a|$\promptiog{i}{o}{,}{g}$)^{\frac{1}{|a|}}$ to better balance actions of various sizes. From these quantities, we build the policy as~$\pi(a|$\promptiog{i}{o}{,}{g}$)= \mathbb{P}^{norm}_{LLM}(a|$\promptiog{i}{o}{,}{g}$)/{Z_{i}^{o,g}}$, where~$Z^{o,g}_i=\sum_{a\in A} \mathbb{P}^{norm}_{LLM}(a|$\promptiog{i}{o}{,}{g}$)$ is the partition function.
Then, we follow the recent textual RL works for sequential decision-making of \citet{carta2023grounding} and \citet{tan2024true} to 
train our LLM agents 
using Proximal Policy Optimization (\ppo)~\citep{schulman2017proximal}. Given a subset of prompt formulations $P_{ppo} \subseteq  P$,
we note~$\pi_{P_{ppo}}$ an LLM agent of parameters $\theta$ trained by minimizing the \ppo loss $PPO_{loss}(\theta,P_{ppo})$ from rollouts using prompt formulations $P_{ppo}$. Each rollout $\tau$ is  obtained for a given 
prompt formulation \formulation{i}$\in P_{ppo}$ uniformly sampled from $P_{ppo}$, and is used by the agent  during training. 

\subsection{Mitigating Prompt Overfitting with Contrastive Learning}
\label{mitig}
After analyzing prompt overfitting within the proposed framework, we add a contrastive loss to bring closer the latent representations $z_\theta(p^{o,g}_i)$ and ~$z_\theta(p^{o,g}_j)$ of the same observation-goal pair $(o,g)$ across prompts \promptiog{i}{o}{,}{g} and \promptiog{j}{o}{,}{g}, compared to latent representations $z_\theta(p^{o',g'}_i)$ of other observations and goals $(o',g')$ of prompt \promptiog{i}{o'}{,}{g'}. That is, we aim to minimize:
\small
\begin{multline}
        C^{(i,j)}(\theta)=\hspace*{-0.5cm}\E\limits_{\substack{(o,g) \sim d^{\pi_{P_{ppo}}} \\ (o',g') \sim d^{\pi_{P_{ppo}}}}} 
        \hspace*{-0.5cm} \max  \left( \Delta(z_\theta(p^{o,g}_i),z_\theta(p^{o,g}_j)) \right.\\ \left.  -\Delta(z_\theta(p^{o,g}_i),z_\theta(p^{o',g'}_i))+1),0 \right)
\end{multline}
\normalsize
\noindent where~$z_\theta(p^{o,g}_i)$ is the latent representation of textual prompt \promptiog{i}{o}{,}{g} produced by prompt formulation~\formulation{i} applied to the observation-goal pair~$(o, g)$,~$\Delta(z,z')=||(z)-(z')||_{2}^{2}$ is the Euclidean distance between two latent representations~$z$ and~$z'$, and~$d^{\pi_{P_{ppo}}}$ is the joint stationary observation-goal distribution of a policy using $P_{ppo}$. In practice, we sample pairs~$(o, g)$ from the \ppo rollouts, supposed to follow~$d^{\pi_{P_{ppo}}}$. 

\noindent Our final loss~$L$ jointly optimizes~$PPO_{loss}$ with the contrastive loss~$C^{(i,j)}(\theta)$ as follows: 
{\small
\begin{eqnarray}   
    \vspace{-0.2cm}
    L(\theta,P_{ppo},P_c)\hspace{-0.1cm}&=&PPO_{loss}(\theta,P_{ppo})\\
    &+&\frac{\alpha}{|P_c|^2} \hspace{-0.1cm}\sum\limits_{P_i ,P_j \in P_c^2 }  \hspace{-0.1cm}C^{(i,j)}(\theta) \nonumber
    \vspace{0.2cm}
\end{eqnarray}
\normalsize}
\noindent where $\alpha$ is a parameter regulating the impact of  $C^{(i,j)}$, $P_{ppo}$ represents the set of prompts used for fine-tuning the LLM policy with the PPO loss, and $P_{c}$ represents the set of prompts used for the contrastive regularization $C^{(i,j)}$. Following~\citep{ni-etal-2022-sentence},~$C^{(i,j)}$ is only applied to the representation of the first token of the prompt. Appendix~\ref{app:token choice} provides implementation details about contrastive regularization.

\section{Experimental Protocol}
In this section, we introduce an evaluation methodology for analyzing the impact of fine-tuning LLMs with RL in textual environments. 
Our experiments aim to address three research questions:

\noindent$\bullet$~\textbf{Q1: Prompt sensitivity:} Are LLMs sensitive to different prompt formulations, and how does this sensitivity influence their ability to generalize across various prompt formulations?

\noindent$\bullet$~\textbf{Q2: State representation:} How do LLMs encode the state space in their hidden representation, and what is the topology of the latent space?

\noindent$\bullet$~\textbf{Q3: Impact of prompt information on action choice:} After fine-tuning with various prompts, on which parts of the prompt does the LLM agent focus for task completion?

\subsection{Environments}
\label{sec: Environments}

We conduct our experiments in two textual environments: 1) BabyAI-Text, a mini-grid environment used in \cite{carta2023grounding} where an agent navigates through limited actions, and 2) the Medium difficulty TWC Environment, proposed in \cite{murugesan2021text} (noted TWC-Medium), targeted to solve household tasks getting a scene description and a list of possible actions.
~These two environments assess complementary skills in terms of semantics: BabyAI-Text requires exploring and understanding the arrangement of objects, whereas TWC-Medium requires common sense knowledge and reasoning. See Appendix~\ref{app:envs} for more details.

\subsection{Prompt Design}
\label{sec: variations}
To use LLMs as policies, we define four distinct prompt formulations to gather several pieces of information from the environment: the goal, possible actions, an inventory and textual observations. The first prompt formulation~\formulation{0} follows a format where pieces of information are separated by line breaks in the following order: possible actions, goal, observation, and inventory.~\formulation{1} is similar to~\formulation{0} but switches the order of the information.~\formulation{2} employs a more rigid syntax with delimiter tags, similar to an XML file. Finally,~\formulation{3} removes all rigidity in syntax and follows a natural language format, paraphrased by a prompt writer.  Examples of~\formulation{0} and~\formulation{2} in TWC-Medium can be found in \figurename~\ref{fig:main GLAM} and detailed descriptions of all prompt formulations are provided in Appendix~\ref{app: strat}.

\subsection{Training and Evaluation Details}
\textbf{Training:} We consider several LLMs including encoder-decoder architectures (Flan-T5 78M, 780M, and 2.7B)~\citep{https://doi.org/10.48550/arxiv.2210.11416} and decoder-only architectures~(GPT-Neo 1.3B, LLama 7B)~\citep{Black2022GPTNeoX20BAO, Touvron2023LLaMAOA}. We present in the main paper the results obtained by Flan-T5 78M and 780M also used in~\citep{carta2023grounding} and GPT-Neo 1.3B. Additional results on other LLMs are presented in Appendix~\ref{app:rq1}.
\move{We consider different fine-tuning scenarios denoted as \scenario.
For brievity, we omit to specify $P_{c}$ if no contrastive regularization is applied. Also, we denote as~\ZS the zero-shot scenario, that corresponds to the pre-trained LLM agent without any fine-tuning (i.e., both $P_{c}$ and $P_{ppo}$ are empty). Three scenarios are mainly considered for fine-tuning LLM agents in our experiments: 
1) the one prompt scenario~\PO where the LLM is fine-tuned by \ppo on prompt \formulation{0} only, 
2) the multiple prompt scenario
~\Pall where the LLM is fine-tuned by \ppo on prompts~\formulation{0} to ~\formulation{3}, and 3) the contrastive scenario 
\contrastive where the LLM is fine-tuned by \ppo with~\formulation{0} only, but using an additional contrastive loss considering prompts~\formulation{0} to~\formulation{3}. 
Fine-tuning is performed in both environments for 500k steps across 5 seeds.}

\noindent\textbf{Evaluation:} We evaluate the zero-shot scenario \ZS and the fine-tuned LLMs (\scenario) to assess the performance on both seen and unseen prompts during training. In addition, for the scenarios~\Pall and \contrastiveO, we define a new prompt formulation~\formulation{4} to analyze generalization to unseen prompts.  \formulation{4} follows a different template with changes to the section's names, reordering, and an additional context tag (details about this prompt in Appendix~\ref{app: strat}).

\subsection{Metrics}

We assess LLM agents around four axes:

\noindent \textbf{$\bullet$ Performance Related to Training Task:} We define a success of an LLM agent in the environment as a case where a trajectory is successful if it reaches its goal~$g$. We deduce two metrics: 1) (\textbf{SR}): the success rate of the agent and 2) ($\overline{\textbf{SR}}$): the mean success rate across all prompt formulations and episodes.

\noindent \textbf{$\bullet$ Exploring Hidden Representations:} To delineate the disparities between prompt formulations, we compute the cosine similarity of their latent representations pre and post fine-tuning. Given a set of observation-goal pairs\small~${\Gamma}=\{(o,g)\}$\normalsize\footnote{In our experiments,~$\Gamma$ is built from rollouts of \POO to infer actions. Similar results were observed with other distributions.}, the similarity between representations using different inputs formatted by a single prompt formulation~\formulation{0} is defined as
\vspace{-0.2cm}
{\begin{equation}
    \vspace{-0.2cm}
    Intra(P_i)=\frac{1}{|\Gamma|^2-|\Gamma|} \hspace*{-0.5cm} \sum_{\substack{(o,g)\in \Gamma \\ \substack{(o',g')\in \Gamma\setminus\{o,g\}} }} \hspace*{-0.5cm} cos\left(z_i^{o,g},z_i^{o',g'}\right). \nonumber 
\end{equation}}
\noindent Besides, we assess the similarity between representations from different prompt formulations as:
\vspace{-0.2cm}
{\begin{equation}
Inter(P_i,P_j)=\frac{1}{|\Gamma|} \sum_{(o,s)\in \Gamma} cos\left(z_i^{o,g},z_j^{o,g}\right).\nonumber 
\end{equation}
}
\noindent Finally, to visualize the topology of the latent space, we utilize UMAP~\citep{McInnes2018} to generate a 2D projection of the hidden state space and visualize its structure.

\noindent \textbf{~$\bullet$ Relevance of Parts of Prompt Information:} We study the relationship between inputs and predictions using gradients attribution methods to generate saliency scores for each element of the input~\citep{Madsen2021PosthocIF}.
For each scenario~$\sigma$ and each prompt formulation~\formulation{i}, we compute the averaged saliency of prompt tokens for every observation-goal pair~$(o,g)$ from~$\Gamma$. Saliency scores are computed using the {\tt Inseq} toolkit~\citep{sarti-etal-2023-inseq}, as the Integrated Gradient of the output  probability of the policy~$\pi(a^*|$\promptiog{i}{o}{,}{g}$)$ with respect to the tokens of the prompt~\promptiog{i}{o}{,}{g}, with~$a^*$ the most likely action regarding the observation-goal $o,g$.
To enhance interpretability, saliency scores for each prompt tokens are finally aggregated depending on the part of the prompt they belong to, among
\{possible actions, goal, observation, inventory\}. We do so after filtering out the top 5\% most significant tokens from each section to eliminate perturbations caused by irrelevant tokens. 

\noindent \textbf{~$\bullet$ Knowledge acquired by the LLM about the environment:} 
\noindent To analyze what the LLM learned about the environment, we measure the accuracy in question-answering~(QA) tasks related to  TWC-Medium that include object counting (TWC OC) and task-related questions (TWC QA). We follow the~\cite{xiang2023language} methodology for constructing the set of questions based on the optimal successful trajectory of TWC-Medium.
For task-related questions, we generate multiple-choice questions regarding the best object from a given set to accomplish a task. For the object counting task, we present to the LLM a trajectory in the environment and prompt it to count the number of objects in a specific location. 
\section{Quantifying overfitting}
\begin{figure*}[!ht]
    \centering
    \includegraphics[width=\linewidth]{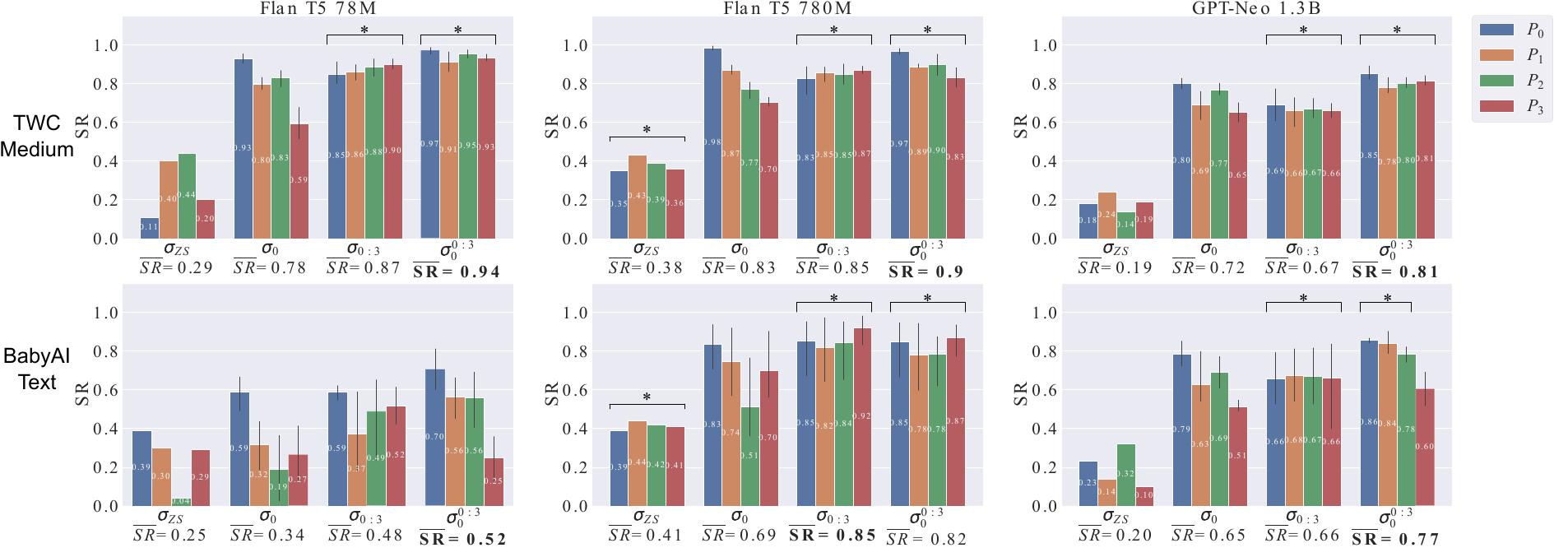}
    
    \caption{
    \textbf{Success Rate (SR)} in BabyAI-Text and TWC-Medium. The x-axis indicates the training scenario, while colors represent the prompt formulation used to format inputs during rollouts. The asterisk (*) indicates instances where the chi-squared test exceeds the homogeneity threshold (p-value > 0.05). Bolded values in $\overline{\textbf{SR}}$ represent the best results in the evaluation. Results show that the LLM exhibits heterogeneous performance when the prompt formulation used during training is changed, with a drop in success rate of up to 30\% in certain scenarios. }
    \label{fig:bar plt sr}
    
\end{figure*}
\label{quantify}
We now focus on the impact of fine-tuning LLMs in interactive environments according to our three main research questions.
We leave the impact of the contrastive learning scenario on task achievement and question answering for Section~\ref{mitig res}.

\subsection{Q1: Prompt sensitivity}
\label{sec:5.1}
We evaluate the LLM performance in solving tasks with different prompt formulations in both environments. Figure~\ref{fig:bar plt sr} shows the SR values for Flan-T5 78M, 780M, and Gpt-Neo 1.3B obtained according to the different evaluation scenarios:~\ZS,~\POO, and~\Pall. Results of training with other formulations and models are available in Appendix~\ref{app:rq1}. 

First, we see that LLM performance in \ZS is 50\% lower than that of models fine-tuned via \ppo. This underscores the necessity for LLM specialization in interactive environments to efficiently achieve goals. Conversely, while the 78M model and GPT-Neo 1.3B  exhibit heterogeneous results across both environments in \ZS, the Flan T5 780M model demonstrates more consistent performance.
 
\noindent Second, fine-tuning the LLM with a single prompt formulation, notably~\formulation{0}, enhances the SR value for the training prompt, revealing that the LLM learns to effectively achieve tasks and capture the dynamics of interactive environments. For more details, the training curves are provided in Appendix~\ref{app:train}. For instance, in the TWC-Medium environment, both the Flan T5 78M and 780M models achieve success rates exceeding 90\% (compared to a maximum of 45\% with \ZS) and approach the 85\% success rate of the GPT-Neo 1.3B (compared to at most 20\% with \ZS). However, a notable decline in performance is observed when using another prompt formulation at test time, with a decrease of over 30\% from using the original~\formulation{0} to the~\formulation{3} variation in TWC-Medium and more than 30\% from~\formulation{0} to~\formulation{2} in BabyAI-Text. This drop outlines prompt overfitting in LLMs. The trend is similar for larger models (Flan T5 2.7B and LLama 7B) as detailed in Appendix~\ref{app:rq1}.
We also observe that, for both \ZS and fine-tuned models, encoder-decoder models outperform decoder-only architectures, even when the decoder-only models have a larger parameter size. 
\label{sec:BabyAISMALL}
\noindent Finally, results obtained with~\Pall indicate that training with all prompts maintains the LLM's performance more consistently across prompt variations. However, the LLM does not achieve the peak of performance observed when trained on a single prompt formulation. This is likely due to the higher difficulty for the LLM to adapt to multiple formulations than to a single formulation.
\noindent By comparing environments, we observe that the Flan T5 78M model struggles in learning appropriate policies in BabyAI-Text. This might be attributed to its exploratory nature and reduced reliance on common sense knowledge, due to the small portion of exploited vocabulary.



\begin{figure*}[!t]
    \centering
    \includegraphics[width=0.80\linewidth,height=100pt]{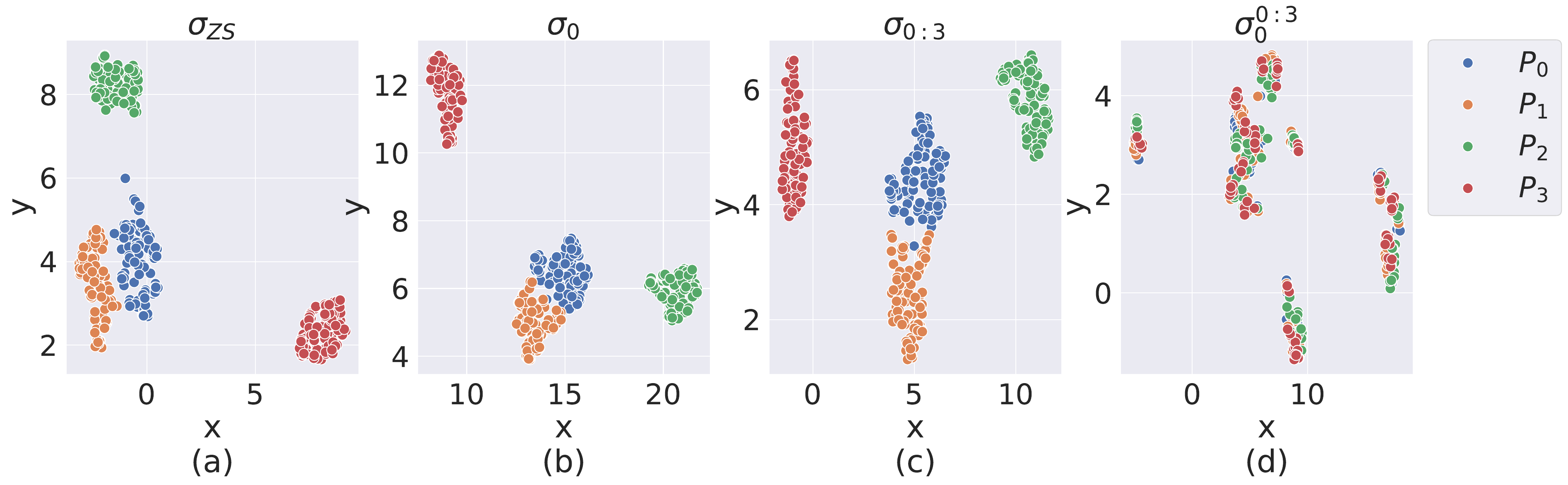}
    
    \caption{\textbf{UMAP visualization of hidden representations in GPT-Neo 1.3B across 100 states of TWC-Medium using four prompt formulations} demonstrating clustering based on prompt formulation over semantic state similarity in both \ZS and fine-tuned models~\POO and~\Pall. Additional results for fine-tuning on other prompts can be found in Appendix~\ref{app:rq3}. } 
    
    \label{fig:umap}
\end{figure*}

\begin{figure*}[!ht]
    \centering
    \includegraphics[width=1\linewidth]{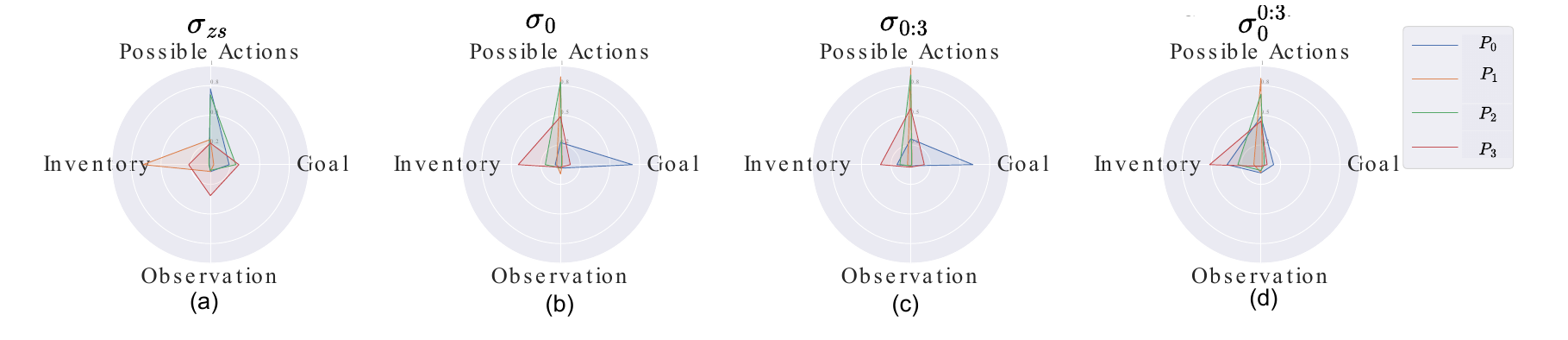}
    
    \caption{\textbf{Saliency maps of the Flan-T5 78M model across scenarios on TWC-Medium}, highlighting various parts of the prompts. The maps show that the LLM focuses on different sections of the prompt depending on the prompt formulation. This variation in focus is linked with performance changes when the prompt formulation is altered. The phenomenon is observed in \ZS, \POO and \Pall scenarios.}
    \label{fig:IG}
    
\end{figure*}

\subsection{Q2: State representation}
\label{sec:5.2}
To further analyze prompt overfitting, we investigate the latent representation of states formatted with different prompt formulations. 
We compare in Table~\ref{tab:distance intra inter} the intra-prompt and inter-prompt similarities to measure the topology of the latent representations of states according to our different scenarios (\ZS,~\PO, and~\Pall), averaged over any pair of prompt formulations used to format inputs. The most striking result is that LLMs tend to cluster prompts formulated with the same prompt formulation ($Intra \simeq 1$), even when they correspond to different goal-observation pairs. In contrast, the same pair formulated with different prompt formulations exhibits low similarity ($Inter< 0.5$).

\begin{table}[ht]
\resizebox{0.5\textwidth}{!}{
\begin{tabular}{|cc|c|c|c|c|}
\hline
\multicolumn{2}{|c|}{Models} & \begin{tabular}[c]{@{}c@{}}\ZS\end{tabular} &~\POO &~\Pall &~\contrastiveO \\ \hline
\multicolumn{1}{|c|}{\multirow{2}{*}{$78M$}} &~$Intra$ & \begin{tabular}[c]{@{}c@{}}0.992 \\~$\pm$ 0.003\end{tabular} & \begin{tabular}[c]{@{}c@{}}0.991 \\~$\pm$ 0.003\end{tabular} & \begin{tabular}[c]{@{}c@{}}0.991 \\~$\pm$ 0.003\end{tabular} & \begin{tabular}[c]{@{}c@{}}0.907 \\~$\pm$ 0.017\end{tabular} \\ \cline{2-6} 
\multicolumn{1}{|c|}{} &~$Inter$ & \begin{tabular}[c]{@{}c@{}}0.376 \\~$\pm$ 0.019\end{tabular} & \begin{tabular}[c]{@{}c@{}}0.382 \\~$\pm$ 0.020\end{tabular} & \begin{tabular}[c]{@{}c@{}}0.371 \\~$\pm$ 0.020\end{tabular} & \begin{tabular}[c]{@{}c@{}}0.806 \\~$\pm$ 0.029\end{tabular} \\ \hline
\multicolumn{1}{|c|}{\multirow{2}{*}{$780M$}} &~$Intra$ & \begin{tabular}[c]{@{}c@{}}0.998\\~$\pm$ 0.001\end{tabular} & \begin{tabular}[c]{@{}c@{}}0.997\\~$\pm$ 0.001\end{tabular} & \begin{tabular}[c]{@{}c@{}}0.998 \\~$\pm$ 0.001\end{tabular} & \begin{tabular}[c]{@{}c@{}}0.939 \\~$\pm$ 0.017\end{tabular} \\ \cline{2-6} 
\multicolumn{1}{|c|}{} &~$Inter$ & \begin{tabular}[c]{@{}c@{}}0.469 \\~$\pm$ 0.462\end{tabular} & \begin{tabular}[c]{@{}c@{}}0.458 \\~$\pm$ 0.449\end{tabular} & \begin{tabular}[c]{@{}c@{}}0.47 \\~$\pm$ 0.461\end{tabular} & \begin{tabular}[c]{@{}c@{}}0.812 \\~$\pm$ 0.06\end{tabular} \\ \hline
\multicolumn{1}{|c|}{\multirow{2}{*}{$1.3 B$}} & \multicolumn{1}{l|}{$Intra$} & \begin{tabular}[c]{@{}c@{}}0.995\\~$\pm$ 0.001\end{tabular} & \begin{tabular}[c]{@{}c@{}}0.994\\~$\pm$ 0.001\end{tabular} & \begin{tabular}[c]{@{}c@{}}0.997 \\~$\pm$ 0.001\end{tabular} & \begin{tabular}[c]{@{}c@{}}0.994\\~$\pm$ 0.017\end{tabular} \\ \cline{2-6} 
\multicolumn{1}{|c|}{} & \multicolumn{1}{l|}{$Inter$} & \begin{tabular}[c]{@{}c@{}}0.552 \\~$\pm$ 0.03\end{tabular} & \begin{tabular}[c]{@{}c@{}}0.539 \\~$\pm$ 0.01\end{tabular} & \begin{tabular}[c]{@{}c@{}}0.501\\~$\pm$ 0.05\end{tabular} & \begin{tabular}[c]{@{}c@{}}0.94\\~$\pm$ 0.009\end{tabular} \\ \hline
\end{tabular}
}

\caption{\label{tab:distance intra inter} \textbf{Inter and intra-similarity} comparison for Flan T5 78M, 780M and GPT-Neo 1.3B models on TWC-Medium on our different scenarios.  For all models, we observe that $Intra$ approaches 1, while $Inter$ is consistently below 0.5. This indicates that the LLMs tend to cluster prompts based on their formulation rather than on content. A similar trend is observed for both~\Pall and~\POO scenarios.
}
\end{table}

The low similarity between prompts and the high similarity within prompts suggest that LLMs capture more about the prompt formulation than about the content itself. This reinforces the previous observation about Q1 highlighting prompt overfitting. 
\noindent Figure~\ref{fig:umap} depicts the embedding of prompts using the UMAP visualization and also corroborates this hypothesis. We see that there is no overlap between clusters of prompt formulations in \ZS, and fine-tuning with~\POO does not change this. Interestingly, using \Pall does not mitigate this, highlighting the need for methods tackling overfitting in both task efficiency (Q1) and state representation (Q2).
\subsection{Q3: Usefulness of Prompt Information}
\label{sec:5.3}
We then analyze which parts of the prompts (goal, possible actions, observation, and inventory) are used by LLMs to predict the best action. \figurename~\ref{fig:IG} depicts the saliency scores of prompt parts obtained using the Integrated Gradients algorithm~\cite{Madsen2021PosthocIF}. We observe that the importance of the parts of the prompt varies regarding the different prompt formulations and scenarios. For instance, in \ZS, LLMs prioritize Inventory when~\formulation{1} is used, whereas switching to~\formulation{0} and~\formulation{2} shifts the focus to Possible Actions, with quite similar saliency scores for both. This can be explained by the fact that \formulation{0} and~\formulation{2} follow the same order of information (see Appendix~\ref{app: strat}), so LLMs find possible actions in the same relative position. Fine-tuning LLMs alters the saliency scores compared to \ZS, with LLMs prioritizing the goal over possible actions when using~\formulation{0}. However, LLMs still focus on different parts of the prompt when changing prompt formulation to the others.
\noindent Using \Pall also results in heterogeneous saliency maps across prompt formulations, showing that LLMs continue to process the prompts differently despite being fine-tuned with multiple prompt formulations. This strengthens the findings of Q2, highlighting the influence of prompt formulation on the behavior of LLMs. Additional results for fine-tuning on other prompts can be found in Appendix~\ref{app:rq3}.

\section{Mitigating Overfitting with Contrastive Learning Regularization}
\label{sec:6}
\label{mitig res}
We now evaluate the impact of the contrastive loss proposed in Section~\ref{mitig} with the aim to help the LLMs focus more on the content in prompts so as to mitigate prompt overfitting.

\noindent\textbf{Performance and prompt overfitting:}
\figurename~\ref{fig:bar plt sr} reports the SR and prompt homogeneity across both environments and three model sizes. In most cases, the $\overline{\textbf{SR}}$ of \contrastiveO surpasses the \ZS, \POO and \Pall (except for the 780M model on BabyAi-Text, where \contrastiveO underperforms by 3\%). This is noticeable since~\contrastiveO learns the matching between prompt formulations while~\Pall learns each prompt independently. In terms of homogeneity, both~\contrastiveO and~\Pall scenarios yield consistent results (marked with an asterisk (*) in \figurename~\ref{fig:bar plt sr}) for all TWC-medium evaluations. However, as mentioned in Section~\ref{sec:BabyAISMALL}, the 78M model struggles in BabyAI-Text with both task-solving and achieving homogeneity across prompts for \contrastiveO and \Pall. For the GPT-Neo 1.3B model, \contrastiveO achieves homogeneity across prompts \formulation{0} to \formulation{2} but its performance drops with \formulation{3}. This is likely because the environment involves extensive exploration, making it difficult to maintain semantic consistency with paraphrased text. Nevertheless, the $\overline{\textbf{SR}}$ for~\contrastiveO remains higher than that of~\Pall.

\noindent We also evaluate in Table~\ref{tab:p4} the generalization capabilities of \contrastiveO compared to~\Pall using the prompt formulation \formulation{4}, which was not seen during training in either scenario.
\begin{table}[t]
\centering
\resizebox{0.45\textwidth}{!}{\begin{tabular}{c|c|c|}
\cline{2-3}
 &~\Pall &~\contrastiveO \\ \hline
\multicolumn{1}{|c|}{78 M} & 0.77~$\pm$ 0.11 {\color[HTML]{CB0000}(3\%)}& 0.92~$\pm$ 0.02 {\color[HTML]{009901} (97\%)} \\ \hline
\multicolumn{1}{|c|}{780 M} & 0.80~$\pm$  0.06 {\color[HTML]{CB0000}(4.7\%)}& 0.86~$\pm$ 0.05 {\color[HTML]{009901} (91\%)} \\ \hline
\multicolumn{1}{|c|}{1.3 B} & 0.66~$\pm$ 0.02 {\color[HTML]{009901} (99\%)}  & 0.76~$\pm$ 0.03 {\color[HTML]{009901} (98\%)} \\ \hline
\end{tabular}}
\vspace{-0.2cm}
\caption{\label{tab:p4}\textbf{ Success Rate (SR) in TWC-Medium} using prompt formulation~\formulation{4} unseen during training. The values in parentheses represent the $\chi^2$ homogeneity test results, red indicates heterogeneity (p-value < 5\%) and green indicates homogeneity (p-value > 5\%).~\contrastive scenario demonstrates superior performance in terms of SR and homogeneity compared to~\Pall across all models.}
\end{table}
The results indicate that the mean success rate on~\formulation{4} is higher for \contrastiveO than for~\Pall, indicating superior generalization capabilities across all three model sizes. To validate these findings, we conduct a $\chi^2$ test to verify the homogeneity of~\formulation{4} results compared to the mean results obtained from the training prompts. The values in parentheses summarize the results, showing that for all \contrastiveO models, the success rates on the unseen prompt~\formulation{4} follow the same distribution as the training success rates. This indicates robust generalization to unseen prompts, in contrast to~\Pall, where the 78M and 780M models do not exhibit the same distribution of results as during training. This finding supports the conclusion that regularization helps mitigate overfitting.

\noindent\textbf{State representation:}
The last column of Table~\ref{tab:distance intra inter} shows a significant increase in the inter-prompt proximity for the 78M, 780M and 1.3B models, confirming our intuition that applying the contrastive loss can help disregard the prompt formulation at the benefit of its content. The scatter plot in~\figurename~\ref{fig:umap}(d) indicates that the topology of the latent space changed. The latent vectors are no longer clustered by formulation and now overlap, suggesting that the model has learned to match between states. Further details regarding distances and visualizations can be found in Appendix~\ref{app:rq2}.

\noindent\textbf{Saliency of prompt information:}
\figurename~\ref{fig:IG}(d) shows that contrastive loss homogenizes saliency maps across prompts, as it impacts their performance. Indeed, we see more homogeneous results in the salient prompt parts across prompt formulations, unlike when training on multiple prompts.

\noindent
\textbf{Knowledge acquired by the LLM about the environment.}
 We measure the accuracy of the LLM in TWC QA and TWC OC datasets before and after fine-tuning over all training scenarios in Table~\ref{tab:6}. 
Before fine-tuning, the LLM struggles to answer environment-related questions, exhibiting poor performance across both QA datasets. Following fine-tuning with~\PO or~\Pall, the accuracy shows only superficial improvement for the TWC QA, and minor enhancements are observed in the TWC OC compared to the~\ZS setting. This indicates that fine-tuning in these scenarios leads to superficial updates. However, fine-tuning with the ~\contrastiveO scenario results in a more substantial improvement compared to other scenarios, with at least a 13\% increase in TWC QA accuracy, and respective gains of 25\%, 35\%, and 43\% for the \Pall, \POO, and \ZS scenarios. These findings highlight that not only enhances robustness to prompt variations but also improves the LLM’s understanding of the environment.

\begin{table}[t]
\centering

\begin{tabular}{c|cc|}
 \cline{2-3} 
 & \multicolumn{1}{c|}{\textbf{TWC QA}} & \textbf{TWC OC} \\ \hline
\multicolumn{1}{|c|}{\ZS} & \multicolumn{1}{c|}{0.4866 *} & 0.0876 ***\\ \hline
\multicolumn{1}{|c|}{\PO} & \multicolumn{1}{c|}{0.4901 *} & 0.1340 ***\\ \hline
\multicolumn{1}{|c|}{\Pall} & \multicolumn{1}{c|}{0.5019 *} & 0.2526 *\\ \hline
\multicolumn{1}{|c|}{\contrastiveO} & \multicolumn{1}{c|}{\textbf{0.6322 }} & \textbf{0.5155 } \\ \hline
\end{tabular}
\vspace{-0.2cm}
\caption{\label{tab:6}~\textbf{Impact on knowledge of the LLM about the environment} on TWC QA and TWC OC datasets. * and *** correspond to the p-value (resp. $<0.05$ and $<0.001$) of Welch’s t-test to compare the performance between \contrastiveO and other scenarios. We observe a significant improvement with \contrastiveO scanerio compared to  \ZS, \POO, and \Pall scenarios across both datasets.}

\end{table}

\noindent Altogether, these results highlight the effectiveness of our contrastive method to mitigate prompt overfitting according to different criteria: 1) the impact of prompt sensitivity on performance, 2) leveraging state description rather than prompt formatting, 3) giving importance to the same information in the prompt and 4) have better knowledge about the environment. It is worth noting that this is achieved with the benefit of better generalization, without sacrificing performance compared to a scenario that leverages multiple prompts simultaneously during fine-tuning.

\section{Conclusion and future work}

In this work, we studied the sensitivity of LLMs to prompt formulation during RL fine-tuning. We introduced an evaluation protocol to assess prompt overfitting, considering Success Rate,  and the role of internal mechanisms such as embeddings and saliency. We evaluated three models of varying sizes (Flan T5 78M and 780M, and GPT-Neo 1.3B) across two environments (BabyAI-text and TWC-medium). The results revealed the impact of prompt overfitting, which increased the model's sensitivity to variations in prompt formulation. To address this, we proposed a contrastive regularization method and demonstrated its effectiveness.
However, this study has limitations.  We focus on solutions where LLM interacts with the world solely through text, and requires detailed descriptions at each step. Evaluating other modalities (like images) will be addressed in future work. 

\section*{Acknowledgments}
Experiments presented in this paper were carried out using the HPC resources of IDRIS under the allocation 2024-[A0151013011] made by GENCI. This work was supported by the European Commission's Horizon Europe Framework Programme under grant No 101070381 (PILLAR-robots) and by PEPR Sharp (ANR-23-PEIA-0008, ANR,
FRANCE 2030).
\section*{Limitations}
Our evaluation relies on fine-tuning LLMs, which is computationally intensive and time-consuming. Furthermore, using RL further slows down the training process due to the need for interactions with the environment and RL optimization. For instance, training a Flan T5 78M model requires four NVIDIA V100-32GB GPUs, while a 780M model necessitates four NVIDIA A100-80GB GPUs and GPT-Neo 1.3B necessitates eight NVIDIA A100-80GB. Consequently, fine-tuning larger models is challenging given our available computational resources.
\section*{Ethical Considerations}
In our research, we investigate the sensitivity of LLMs to prompts and propose solutions to mitigate this issue. We believe that our efforts to reduce sensitivity and align LLM outputs with human intentions will be beneficial for the application of LLMs in real-world tasks, and represent a step forward for the implementation of LLMs in robotics tasks and beyond.
\bibliography{llm_skills}

\newpage

\appendix

\section*{Appendix}

\begin{figure*}[ht]
    \centering
    \includegraphics[width=\linewidth]{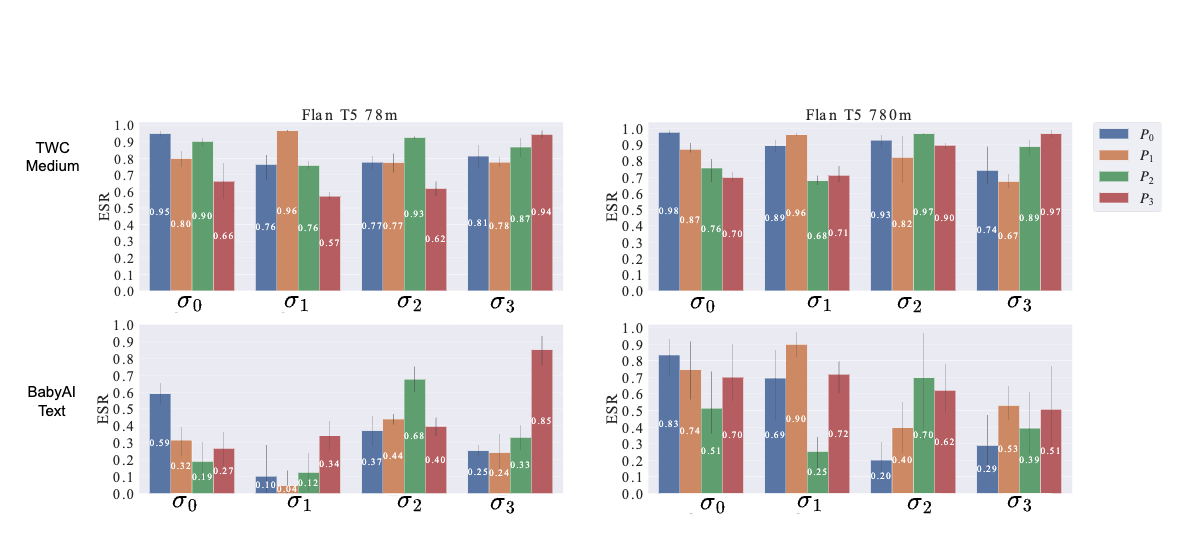}
    \caption{\textbf{Complementary results of success rates} for both the 78M and 780M models in two environments across prompt formulations~\formulation{0} to~\formulation{3}.}
    \label{fig:sr Appendix}
\end{figure*}

\section{Experimental setup}
\label{app:envs}
\subsection{Environments}
The BabyAI-Text environment~\citep{carta2023grounding} encapsulates BabyAI, a simple mini-grid environment where an agent navigates and interacts with objects through a limited action space of six commands: \textit{turn left, turn right, go forward, pick up, drop, and toggle}. BabyAI-Text describes each observation with sentences instead of using a symbolic representation. A textual description consists of a list of template descriptions with the following structure:
\begin{itemize}
    \item "You see a <object> <location>" if the object is a key, a ball, a box or a wall
    \item "You see a(n) open/closed door <location>", if the agent sees a door.
    \item "You carry a <object>", if the agent carries an object.
\end{itemize}

The TWC environment~\citep{murugesan2021text} is notably more complex than BabyAI-Text regarding objects and actions and more amenable to common sense knowledge. TWC agents should achieve a series of household tasks, such as "picking up an apple and placing it in an appropriate location". The agent receives a description of a scene and a list of plausible actions. It must then decide which action to take given the current game state. Successful actions are rewarded with points.

TWC games are categorized into easy, medium, and hard difficulties. As difficulty increases, the number of target objects and rooms to clean up also increases, as detailed in Table~\ref{tab:lvl}.

\begin{table}[htbp]
\centering
\begin{tabular}{l|r|r|r}
                & \multicolumn{1}{l|}{\textbf{Objects}} & \multicolumn{1}{l|}{\textbf{Targets}} & \multicolumn{1}{l}{\textbf{Rooms}} \\ \hline
\textbf{Easy}   & 1                                     & 1                                     & 1                                  \\
\textbf{Medium} & 2-3                                   & 1-3                                   & 1                                  \\
\textbf{Hard}   & 6-7                                   & 5-7                                   & 1-2                               
\end{tabular}
\caption{ \label{tab:lvl}Number of objects, target objects and rooms in TWC games per difficulty level.} 
\end{table}

To choose a difficulty level, we first conducted a \ZS evaluation on TWC-Easy using different prompt formulations. The results, summarized in Table~\ref{tab:twc easy}, indicate that the LLM does not encounter significant difficulties in solving tasks in \ZS. This is why we also performed training and evaluation on TWC-Medium, where the LLM struggles in \ZS, to better analyze its performance and adaptation.

\begin{table}[htbp]
\centering
\begin{tabular}{c|cccc}
\hline
&~\formulation{0}   &~\formulation{1}  &~\formulation{2}   &~\formulation{3}   \\ \hline
\begin{tabular}[c]{@{}c@{}}78M\end{tabular} & 0.73 & 0.81 & 0.75 & 0.83 \\ \hline
\begin{tabular}[c]{@{}c@{}} 780M\end{tabular} & 0.83 & 0.9 & 0.86 & 0.88 \\ 
\hline
\end{tabular}
\caption{\label{tab:twc easy} LLM evaluation in zero-shot on TWC-Easy}
\end{table}
\subsection{Grounding Evaluation}
\label{app:grounding}
In this section, we provide details on the dataset used for grounding evaluation, following the methodology of~\citep{xiang2023language}. The evaluation consists of two tasks:

(1) QA task, where the model is asked to identify the relevant object needed to complete a household activity. 
Example: \textit{"Question: To wash clothes, a possibly related item could be. Possible answer: ["highlighter", "crackers", "laundry detergent", "cupcake"] Answer:}

(2) Object counting task, where the model must determine the number of objects in a specific location. Example: \textit{"you opened a cooking pot and grabbed an apple. Next, you pulled out a dish bowl and scrubbed another apple. He found a bookshelf and put the first apple on it. Then, you opened a clothes pile and washed it before putting it on the same bookshelf. He grabbed another dish bowl and plate and put the cutlets on the bookshelf. He moved the clothes pile, grabbed the first apple and moved it back to its original spot on the bookshelf. Finally, you put the second dish bowl on the bookshelf. How many items are there on the bookshelf?"}
\subsection{Models and Prompt Variations}
\label{app: strat}
In this section, we detail the prompt formulations and provide examples of trajectories performed by the agent.

\begin{figure}[htbp]
    \centering
    \includegraphics[width=1.05\linewidth]{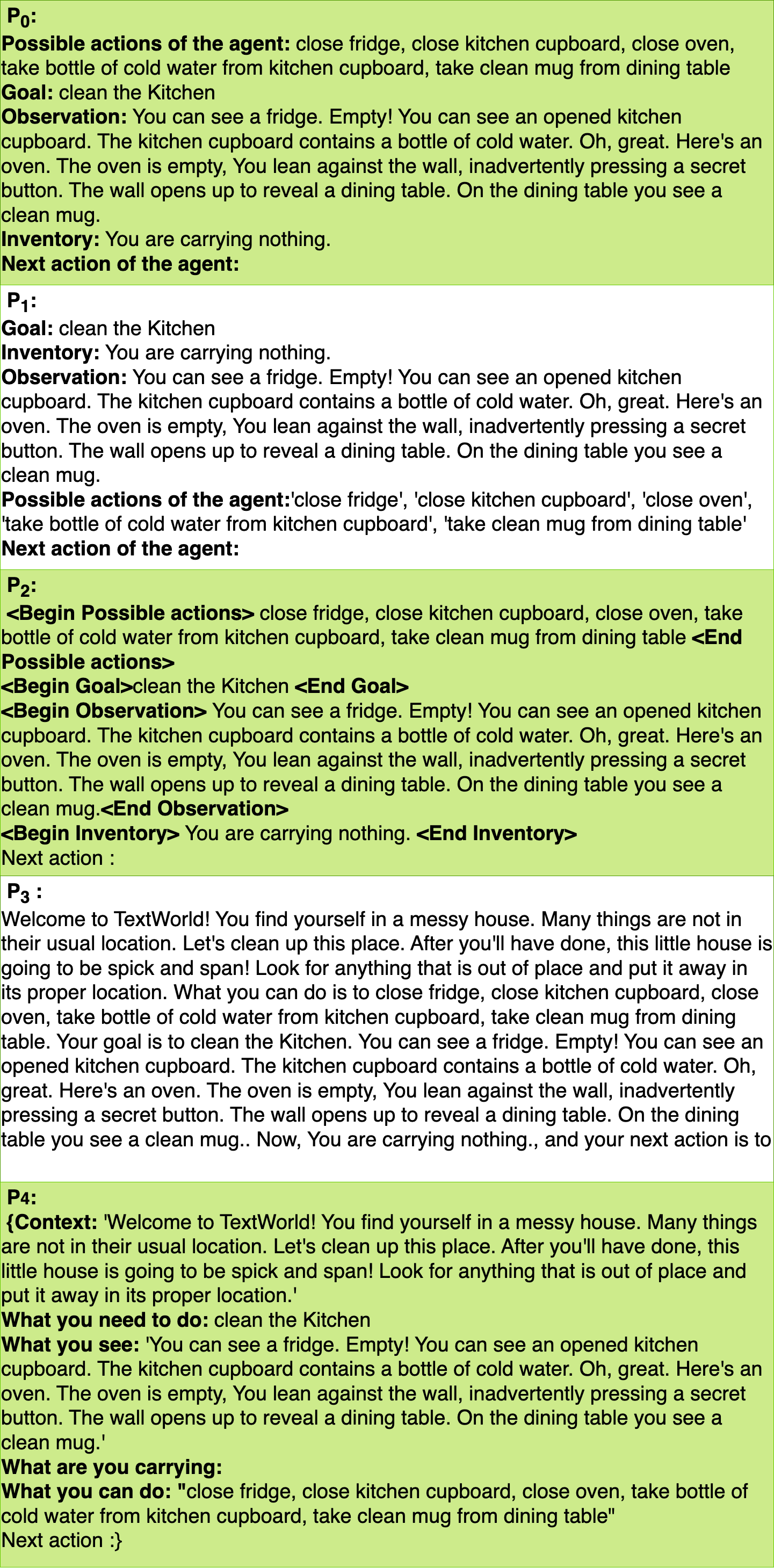}
    \caption{Example of a state described using different prompt formulations in TWC-Medium.}
    \label{fig:Exemple}
\end{figure}

\begin{figure}[htbp]
    \centering
    \includegraphics[width=1.05\linewidth]{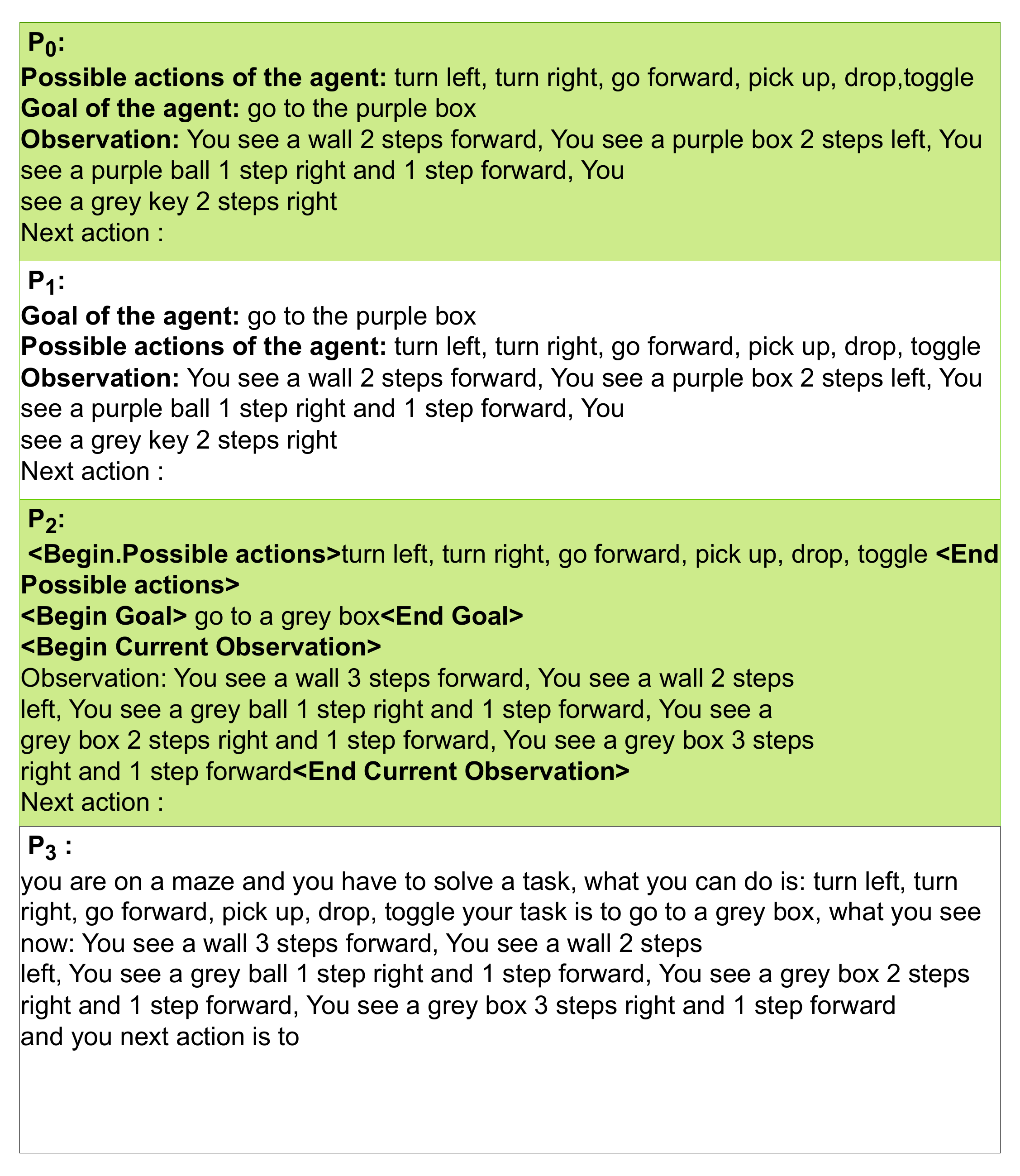}
    \caption{Example of a state described using different prompt formulations in BabyAi-Text.}
    \label{fig:Exemple 2}
\end{figure}

Figure~\ref{fig:Exemple} shows an example of the initial state \(s^1\) in TWC-Medium, formatted with different prompt formulations. Similarly, \figurename~\ref{fig:Exemple 2} presents the same for BabyAI-Text.
\subsection{Training}
\label{app:train}
\begin{figure}[H]
    \centering
    \includegraphics[width=\linewidth]{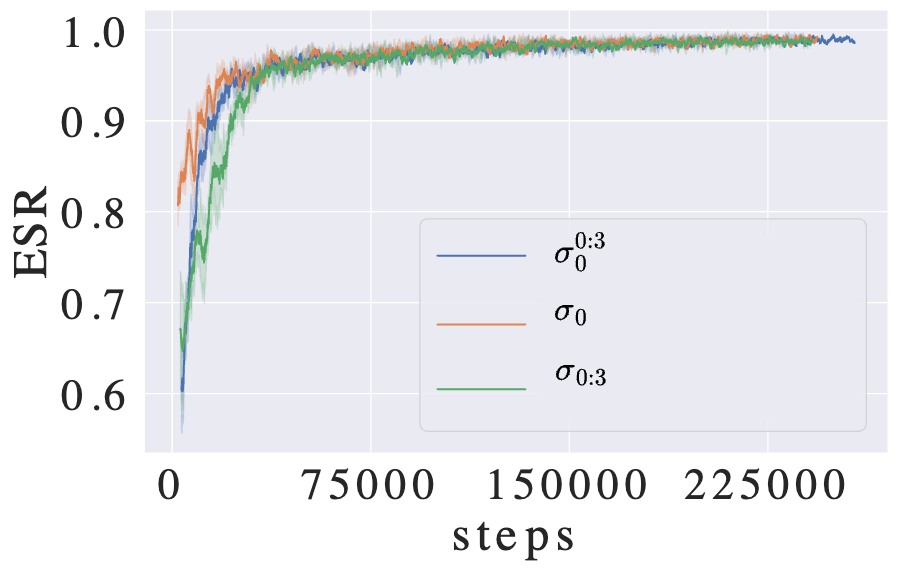}
    \caption{\textbf{Evolution of the Success Rate (SR)} during training in TWC-Medium.}
    \label{fig:training curve}
\end{figure}

In this section, we present success rate curves for models trained on \POO, \Pall, and \contrastiveO(see \figurename~\ref{fig:training curve}). All training scanrios converge, exceeding a 90\% success rate. This indicates that the policy effectively learned to solve the required tasks. The models are trained for an identical number of steps to ensure a fair performance comparison.
\section{Quantifying overfitting} 
\label{app:quantif}
In this section, we present additional results on the three questions studied in the main paper.
\subsection{Q1: Prompt sensitivity}
\label{app:rq1}
First, we perform the same success rates analysis for LLMs trained with different prompt formulations in two environments. Figure~\ref{fig:sr Appendix} displays these results. Prompt overfitting is also evident when fine-tuning with~$P_1$,~$P_2$, and~$P_3$, further supporting our findings. Additionally, we note that the 78M model struggles with BabyAI-Text but shows excellent performance when trained on~$P_3$, indicating that the LLM can better adapt to certain prompt formulations.

We also evaluate larger models like Flan T5 2.7B~\citep{https://doi.org/10.48550/arxiv.2210.11416} and LLama 7B~\citep{Touvron2023LLaMAOA}.
Table~\ref{tab:llama} summarizes the results, demonstrating that prompt overfitting is also present in larger models. Due to the computational cost of fine-tuning such large LLMs, this evaluation was performed exclusively on the TWC environment with \POO scenario.
\begin{table}[ht]
\resizebox{0.45\textwidth}{!}{\begin{tabular}{|l|c|c|c|c|}
\hline
Models &~\formulation{0} &~\formulation{1} &~\formulation{2} &~\formulation{3} \\ \hline
Llama 7B fine-tuned with \POO & 0.9 & 0.72 & 0.75 & 0.79 \\ \hline
Flan T5 XL 2.7B fine-tuned on \POO & 0.93 & 0.89 & 0.84 & 0.74 \\ \hline
\end{tabular}}
\caption{\label{tab:llama} Success Rate (SR) for LLaMA and T5-XL fine-tuned on a single prompt in the TWC environment shows a similar trend of prompt overfitting, even in larger models. }
\end{table}
\begin{figure}[H]
    \includegraphics[width=\linewidth]{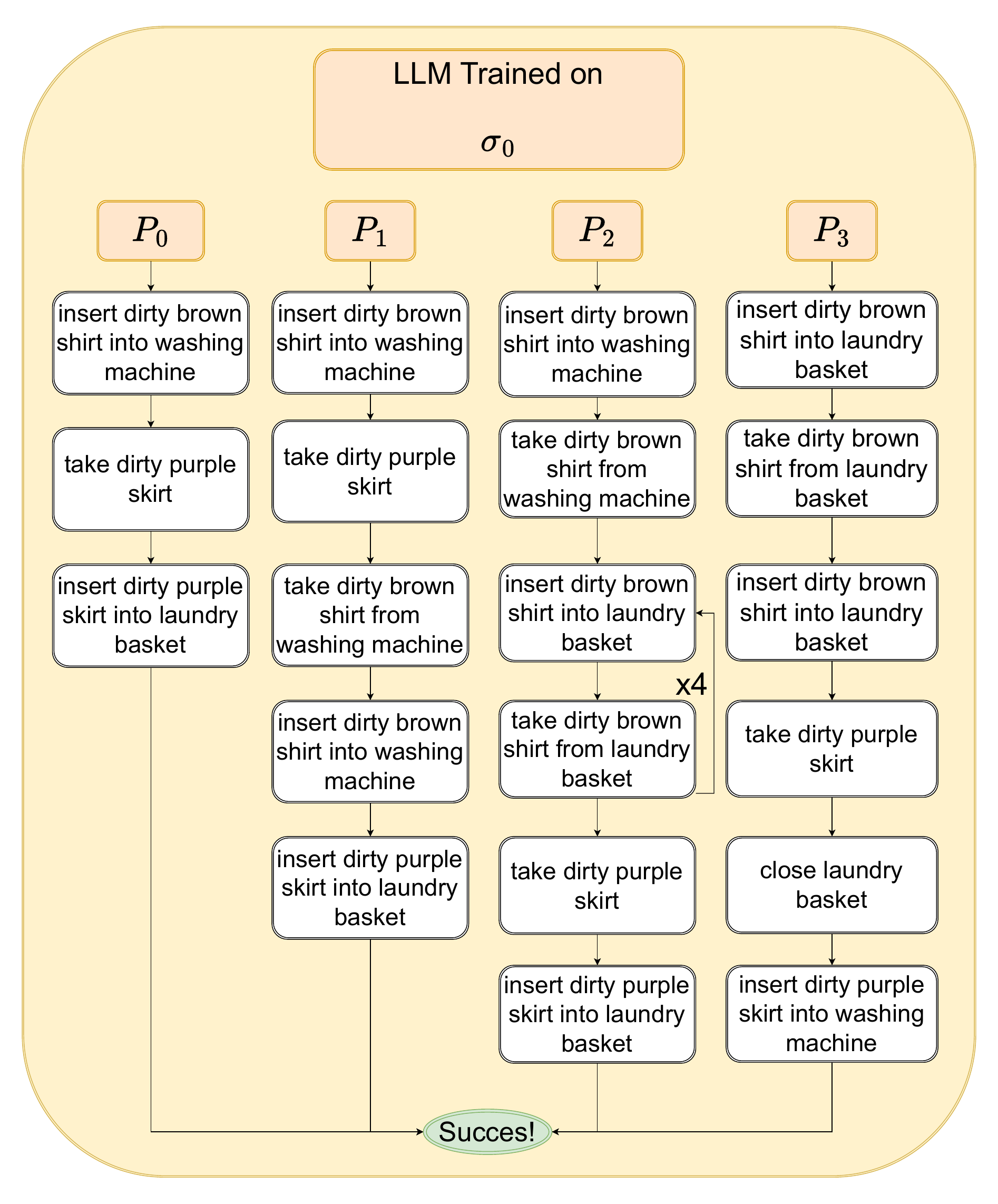}
    \caption{\textbf{Trajectories of the 780M Agent Trained with}~\POO and queried using prompts~\formulation{0} to~\formulation{0}: each vertical line represents a trajectory with the prompt formatted using a specific formulation~\formulation{}, and each step denotes the successive actions taken by the LLM until the goal is achieved.}
    \label{fig:trajectory}
\end{figure}

Indeed, fine-tuning with a single prompt formulation yields good results when evaluated on that same prompt, but there is a significant drop in performance when evaluated on different prompt formulations.
Another key point is the episode length of successful episodes when changing prompt formulations. Figure~\ref{fig:trajectory} shows an example of four trajectories of the 780 model trained with~\POO and evaluated across \formulation{0}, \formulation{1}, \formulation{2} and\formulation{3}. The minimal number of actions is observed when formatted with the formulation used during training, i.e.~\formulation{0}. By contrast, changing the prompt formulation results in irrelevant actions. For instance, when formatted with~$P_2$, the LLM loops four times between two actions before moving to the correct actions to achieve the goal. This further corroborates the impact of prompt overfitting.

\begin{figure*}[t]
    \centering
    \includegraphics[width=\linewidth]{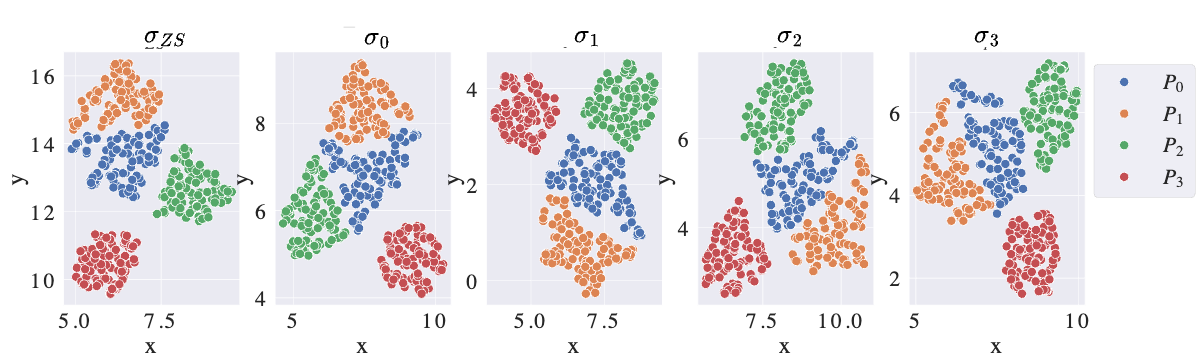}
    \caption{\textbf{UMAP visualization of the hidden representations of Flan T5 780M} with \ZS and fine-tuning with ~\Pf{0},~\Pf{1},~\Pf{2} and~\Pf{3} on TWC-Medium.}
    \label{fig:scatter twc large}
\end{figure*}

\begin{figure*}[t]
    \centering
    \includegraphics[width=\linewidth]{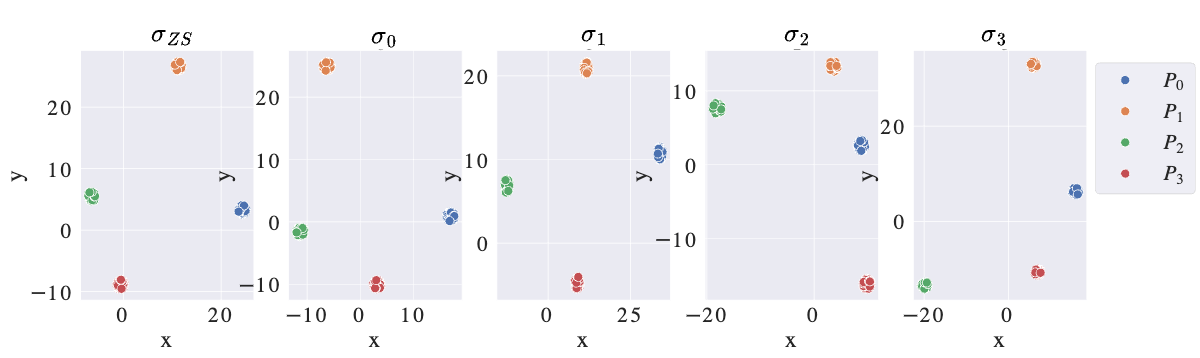}
    \caption{\textbf{UMAP visualization of the hidden representations of Flan T5 78M} with \ZS and fine-tuning with ~\formulation{0},~\formulation{1},~\formulation{2} and~\formulation{3} on BabyAI-Text.}
    \label{fig:scatter baby small}
\end{figure*}

\subsection{Q2: State representation}
\label{app:rq2}
While Table~\ref{tab:distance intra inter} displays the mean intra and inter-prompt similarities for LLMs in \ZS, \POO, \Pall, and with \contrastive, Table~\ref{tab:intra inter Appendix} provides an individual breakdown of these similarities.
\begin{table}[H]
\resizebox{0.5\textwidth}{!}{
\begin{tabular}{cccccc}
\hline\hline
\multicolumn{1}{l}{Models}                 & similarity         & \begin{tabular}[c]{@{}c@{}}\ZS\end{tabular} & \begin{tabular}[c]{@{}c@{}}\POO\end{tabular} & \begin{tabular}[c]{@{}c@{}}\Pall\end{tabular} & \begin{tabular}[c]{@{}c@{}}\contrastiveO\end{tabular} \\ \hline \hline
\multicolumn{1}{c|}{\multirow{7}{*}{78M}}  &~$Intra(P_0)$     &~$0.988$                                              &~$0.987$                                                  &~$0.987$                                                    &~$0.938$                                                  \\
\multicolumn{1}{c|}{}                      &~$Intra(P_1)$     &~$0.989$                                              &~$0.988$                                                  &~$0.988$                                                    &~$0.834$                                                  \\
\multicolumn{1}{c|}{}                      &~$Intra(P_2)$     &~$0.999$                                              &~$0.999$                                                  &~$0.999$                                                    &~$0.999$                                                  \\
\multicolumn{1}{c|}{}                      &~$Intra(P_3)$     &~$0.993$                                              &~$0.992$                                                  &~$0.993$                                                    &~$0.858$                                                  \\ \cline{2-6} 
\multicolumn{1}{c|}{}                      &~$Inter(P_0, P_1)$ &~$0.384$                                              &~$0.388$                                                  &~$0.390$                                                    &~$0.853$                                                  \\
\multicolumn{1}{c|}{}                      &~$Inter(P_0, P_2)$ &~$0.319$                                              &~$0.310$                                                  &~$0.295$                                                    &~$0.726$                                                  \\
\multicolumn{1}{c|}{}                      &~$Inter(P_0, P_3)$ &~$0.427$                                              &~$0.448$                                                  &~$0.429$                                                    &~$0.823$                                                  \\ \hline \hline
\multicolumn{1}{c|}{\multirow{7}{*}{780M}} &~$Intra(P_0)$     &~$0.998$                                              &~$0.987$                                                  &~$0.998$                                                    &~$0.938$                                                  \\
\multicolumn{1}{c|}{}                      &~$Intra(P_1)$     &~$0.998$                                              &~$0.988$                                                  &~$0.997$                                                    &~$0.834$                                                  \\
\multicolumn{1}{c|}{}                      &~$Intra(P_2)$     &~$0.999$                                              &~$0.999$                                                  &~$0.999$                                                    &~$0.999$                                                  \\
\multicolumn{1}{c|}{}                      &~$Intra(P_3)$     &~$0.999$                                              &~$0.992$                                                  &~$0.999$                                                    &~$0.858$                                                  \\ \cline{2-6} 
\multicolumn{1}{c|}{}                      &~$Inter(P_0.P_1)$ &~$0.623$                                              &~$0.388$                                                  &~$0.624$                                                    &~$0.853$                                                  \\
\multicolumn{1}{c|}{}                      &~$Inter(P_0.P_2)$ &~$0.165$                                              &~$0.310$                                                  &~$0.164$                                                    &~$0.726$                                                  \\
\multicolumn{1}{c|}{}                      &~$Inter(P_0.P_3)$ &~$0.619$                                              &~$0.448$                                                  &~$0.622$                                                    &~$0.823$                                                  \\ \hline\hline
\end{tabular} }

\caption{\textbf{Detailed inter and intra-similarity for Flan T5 78M and 780M models} on TWC-Medium. We observe the same clustering behavior depending on the prompt formulation when analyzing each prompt independently.  \label{tab:intra inter Appendix}}
\end{table}

Besides, we further examine the UMAP visualization of models trained with  ~\Pf{0},~\Pf{1},~\Pf{2} and~\Pf{3}, for TWC-Medium to observe potential cluster separation based on prompt formulation rather than semantic content. Results in \figurename~\ref{fig:scatter twc large} reveal distinct clusters corresponding to different prompt formulations, aligning with the findings presented in Table~\ref{tab:intra inter Appendix}. Similarly, UMAP visualization with the 78M model in BabyAI-Text (see \figurename~\ref{fig:scatter baby small}) underscores the persistence of prompt overfitting across different model sizes and environments. Notably, the clustering tendency appears more pronounced in BabyAI-Text, which may elucidate the challenges encountered in implementing the contrastive solution in the 78M model variant on BabyAI-Text.

\subsection{Q3: Relevance of Prompt Information in Decision-Making}
\label{app:rq3}
\begin{figure}[htbp]
    \centering
    \includegraphics[width=\linewidth]{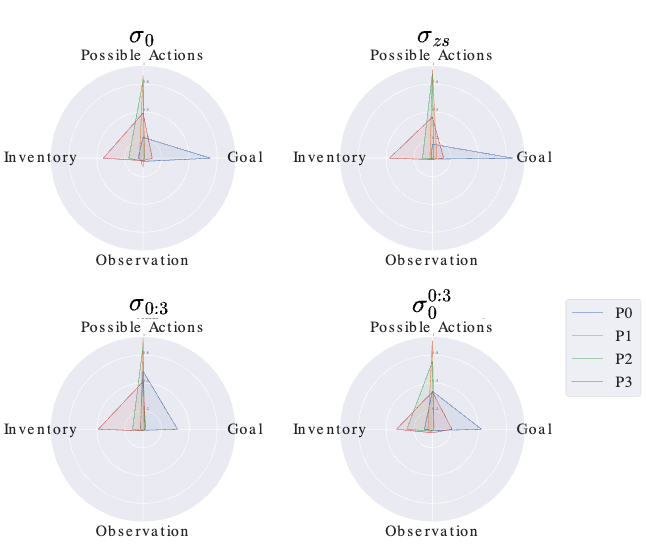}
    \caption{\textbf{Complementary results of Saliency maps} of different parts of prompts for fine-tuned T5 78M models on~\Pf{0} to~\Pf{3}.}
    \label{fig:heatmap Appendix}
\end{figure}

For evaluating the relevance of subparts information of the input prompt, we assess models fine-tuned with  ~\Pf{0},~\Pf{1},~\Pf{2} and~\Pf{3}. Results in \figurename~\ref{fig:heatmap Appendix}, show a variability in the importance of prompt parts across different prompt formulations, even after fine-tuning the LLM. 

\begin{figure}[htbp]
    \includegraphics[width=\linewidth]{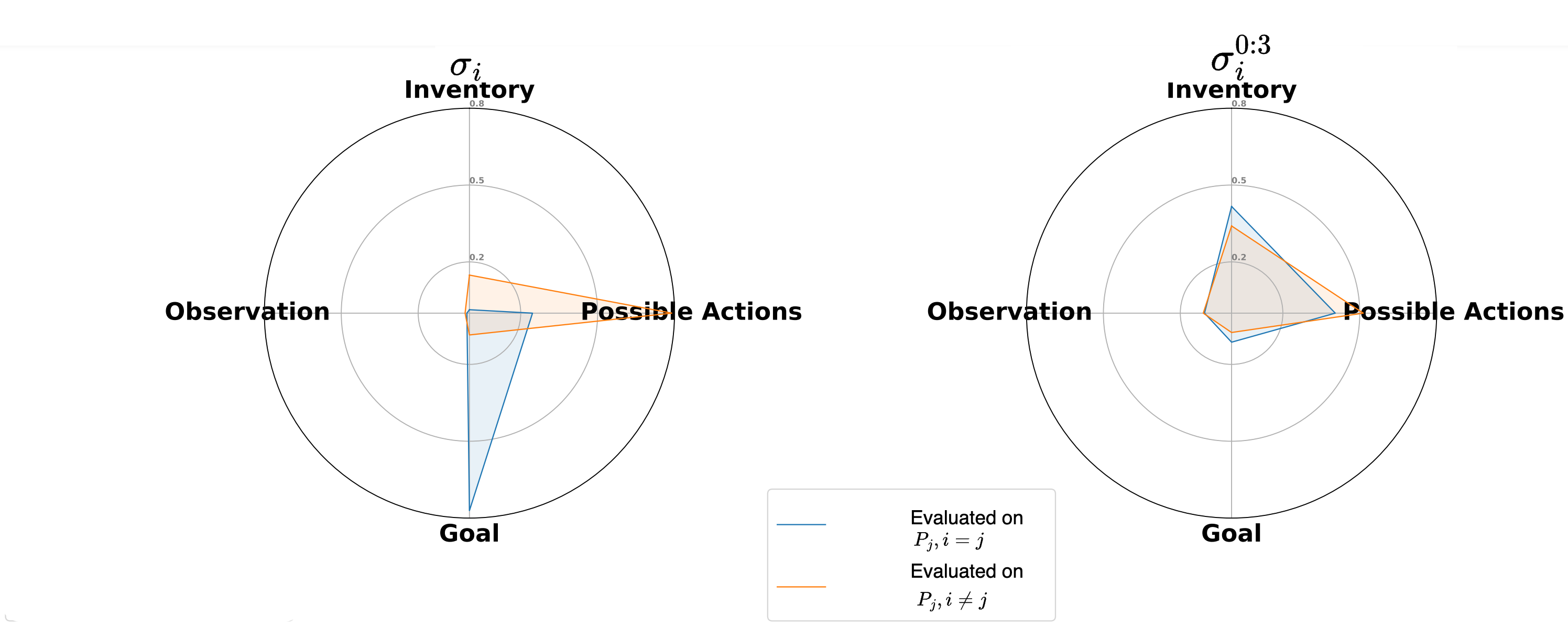}
    \caption{\textbf{Saliency maps} when~$P_i=P_j$ and when~$P_i \neq P_j$.}
    \label{fig:pij}
\end{figure}

Figure~\ref{fig:pij} summarizes outcomes based on two types of prompt formulations: when the formulation~\formulation{0} used during fine-tuning is equal to the evaluation formulation~$P_j$ ($P_i=P_j$) and when~\PO differs from~$P_j$.

Notably, the contrastive regularization approach is more homogeneous, indicating consistent behavior of the LLM across various prompt formulations.
\section{Mitigating Overfitting with Contrastive Learning Regularization}
\label{app:token choice}
\begin{figure*}[t]
    \centering
    \includegraphics[width=\linewidth]{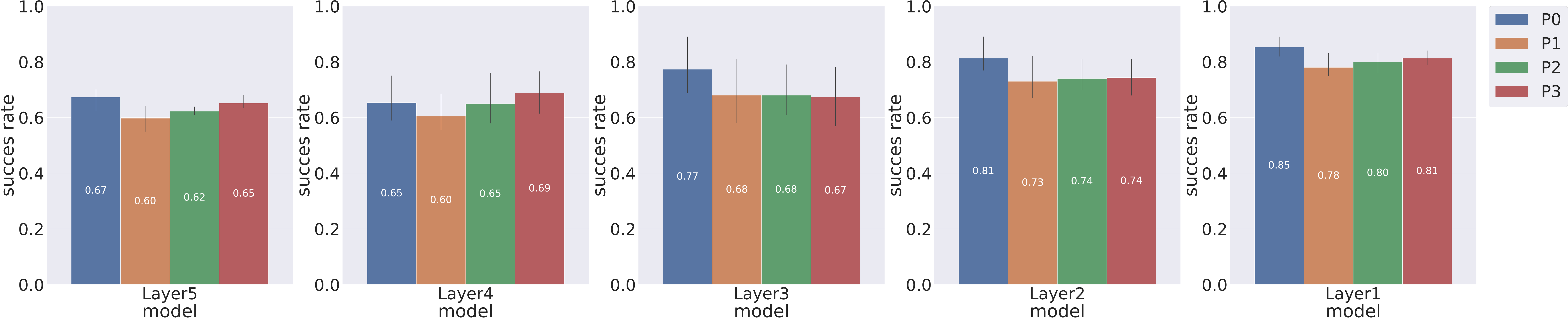}
    \caption{Comparison of different layer choices for applying contrastive regularization on GPT-Neo 1.3B reveals that earlier layers provide better performance. In contrast, performance declines as one moves toward the later layers of the network.}
    \label{fig:token}
\end{figure*}
In this section, we clarify the rationale behind the choice of the regularization token for applying the contrastive method and its adaptation to both encoder-decoder and decoder-only architectures. Two pooling methods can be employed: mean pooling of token embeddings or using the first token embedding. We conducted an evaluation (see Table~\ref{tab:meanfirst}) on both methods and observed that applying contrastive regularization to the mean embeddings does not significantly affect performance, whereas using the first token embedding yields better regularization. We interpret this as the contrastive regularization applied to the first token influencing the embeddings of other tokens through attention mechanisms.
\begin{table}[ht]
\resizebox{0.48\textwidth}{!}{
\begin{tabular}{l|l|l|l|l|}
\cline{2-5}
 &~\formulation{0} &~\formulation{1} &~\formulation{2} &~\formulation{3} \\ \hline
\multicolumn{1}{|l|}{Contrastive Mean Token} & \cellcolor[HTML]{FFFDFA}{\color[HTML]{333333} 0.85} & \cellcolor[HTML]{FFFDFA}{\color[HTML]{333333} 0.71} & \cellcolor[HTML]{FFFDFA}{\color[HTML]{333333} 0.82} & \cellcolor[HTML]{FFFDFA}{\color[HTML]{333333} 0.66} \\ \hline
\multicolumn{1}{|l|}{Contrastive First Token} & \cellcolor[HTML]{FFFDFA}{\color[HTML]{2C3A4A} \textbf{0.97}} & \cellcolor[HTML]{FFFDFA}{\color[HTML]{2C3A4A} \textbf{0.91}} & \cellcolor[HTML]{FFFDFA}{\color[HTML]{2C3A4A} \textbf{0.95}} & \cellcolor[HTML]{FFFDFA}{\color[HTML]{2C3A4A} \textbf{0.93}} \\ \hline
\end{tabular}}
\caption{\label{tab:meanfirst} Comparison of the effect of contrastive regularization applied to the first token versus mean token embeddings shows that using the first token provides better regularization compared to mean token embeddings.}
\end{table}

\begin{figure*}[!ht]
    \centering
    \includegraphics[width=\linewidth]{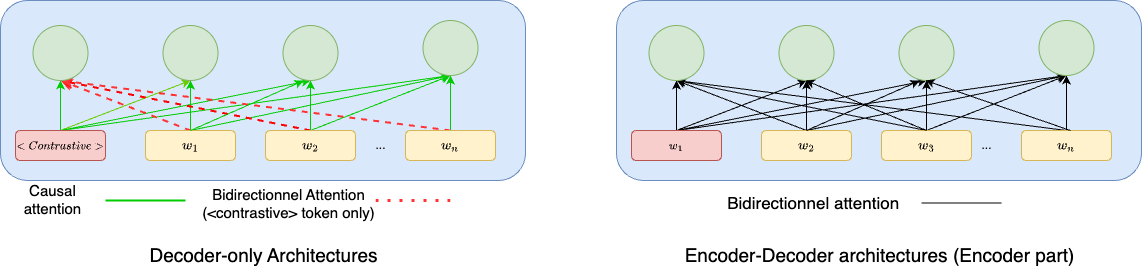}
    \caption{\label{fig:regularization}Selection of a regularization token for encoder-decoder and decoder-only architectures. The regularization token is highlighted in red. In decoder-only architectures, bidirectional attention is permitted for the regularization token, enabling it to access the entire prompt and encode the semantics.}
    
\end{figure*}

For encoder-decoder architectures, we apply regularization to the encoding of the first token in the output of the encoders, following the work of ~\citep{ni-etal-2022-sentence}. In contrast, for decoder-only architectures, the causal nature of these networks means that directly applying contrastive regularization to the first token can diminish the contrastive loss, as the first token does not have access to future tokens. To address this issue, we introduce a new token, <contrastive>, positioned at the beginning of the network, which allows bidirectional attention exclusively for this token. This approach enables the <contrastive> token to access the entire prompt, thereby influencing the embeddings of future tokens. \figurename~\ref{fig:regularization} summarizes the method.
Additionally, we examined which layers are most effective for applying contrastive regularization. Empirical tests conducted on the first five layers of the networks evaluated the impact of contrastive regularization on performance. The results, presented in Figure \ref{fig:token}, indicate that applying the regularization at the beginning of the network yields superior performance, while subsequent layers can still be utilized for task learning with the regularization applied. Additionally, we found in our experiment, that setting $\alpha = 0.5$ in Equation 2 provides a better balance between the PPO loss and the contrastive loss.

\label{app:Contrastive}
Table~\ref{tab:khi2 table} provides complementary results on the differences between training on a single prompt formulation, all prompts, and the contrastive learning scenario, by displaying the mean SR and chi-squared results for the different models and training scenarios. 

\begin{table}[H]
\resizebox{0.5\textwidth}{!}{
\begin{tabular}{|c|c|c|c|c|c|c|}
\hline
ENV                           & Model                  & Metrics & \ZS                                                            & \PO                                                                         & \Pall                                                        & \contrastiveO                                                              \\ \hline
\multirow{4}{*}{\textbf{TWC}} & \multirow{2}{*}{78 M}  & SR     & \begin{tabular}[c]{@{}c@{}}0.28\\~$\pm$0.13\end{tabular}               & \begin{tabular}[c]{@{}c@{}}0.80 \\~$\pm$ 0.12\end{tabular}                 & \begin{tabular}[c]{@{}c@{}}0.88 \\~$\pm$ 0.04\end{tabular} & \begin{tabular}[c]{@{}c@{}}0.94 \\~$\pm$0.03\end{tabular}                \\ \cline{3-7} 
                              &                        &~$\chi^2$ & \begin{tabular}[c]{@{}c@{}}7x$10^-4$\\~$\pm$6,6x$10^{-4}$\end{tabular} & \begin{tabular}[c]{@{}c@{}}1,49x$10^-4$\\~$\pm$2x$10^{-4}$\end{tabular}    & \begin{tabular}[c]{@{}c@{}}0.99\\~$\pm$0.01\end{tabular}   & \begin{tabular}[c]{@{}c@{}}0.99\\~$\pm$0.01\end{tabular}                 \\ \cline{2-7} 
                              & \multirow{2}{*}{780 M} & SR     & \begin{tabular}[c]{@{}c@{}}0.38 \\~$\pm$ 0.03\end{tabular}             & \begin{tabular}[c]{@{}c@{}}0.83 \\~$\pm$ 0.11\end{tabular}                 & \begin{tabular}[c]{@{}c@{}}0.87 \\~$\pm$ 0.03\end{tabular} & \begin{tabular}[c]{@{}c@{}}0.89 \\~$\pm$ 0.05\end{tabular}               \\ \cline{3-7} 
                              &                        &~$\chi^2$ & \begin{tabular}[c]{@{}c@{}}0.99\\~$\pm$0.01\end{tabular}               & \begin{tabular}[c]{@{}c@{}}4,5x$10^{-2}$\\~$\pm$6$\times10^{-3}$\end{tabular}    & \begin{tabular}[c]{@{}c@{}}0.99\\~$\pm$0.01\end{tabular}   & \begin{tabular}[c]{@{}c@{}}0.98\\~$\pm$0.01\end{tabular}                 \\ \hline
\multirow{4}{*}{\textbf{BabyAI}} & \multirow{2}{*}{78 M}  & SR     & \begin{tabular}[c]{@{}c@{}}0.25 \\~$\pm$ 0.13\end{tabular}             & \begin{tabular}[c]{@{}c@{}}0.38 \\~$\pm$ 0.23\end{tabular}                 & \begin{tabular}[c]{@{}c@{}}0.49\\~$\pm$ 0.17\end{tabular}  & \begin{tabular}[c]{@{}c@{}}0.52 \\~$\pm$ 0.21\end{tabular}               \\ \cline{3-7} 
                              &                        &~$\chi^2$ & \begin{tabular}[c]{@{}c@{}}1x$10^{-4}$\\~$\pm$$10^{-4}$\end{tabular}   & \begin{tabular}[c]{@{}c@{}}3,09x$10^{-4}$\\~$\pm$2x$10^{-4}$\end{tabular}  & \begin{tabular}[c]{@{}c@{}}0.531\\~$\pm$0.21\end{tabular}  & \begin{tabular}[c]{@{}c@{}}3,1x$10^{-3}$\\~$\pm$4x$10^{-3}$\end{tabular} \\ \cline{2-7} 
                              & \multirow{2}{*}{780 M} & SR     & \begin{tabular}[c]{@{}c@{}}0.41 \\~$\pm$ 0.01\end{tabular}             & \begin{tabular}[c]{@{}c@{}}0.63 \\~$\pm$ 0.24\end{tabular}                 & \begin{tabular}[c]{@{}c@{}}0.85 \\~$\pm$ 0.12\end{tabular} & \begin{tabular}[c]{@{}c@{}}0.82 \\~$\pm$ 0.12\end{tabular}               \\ \cline{3-7} 
                              &                        &~$\chi^2$ & \begin{tabular}[c]{@{}c@{}}0.99\\~$\pm$0.01\end{tabular}               & \begin{tabular}[c]{@{}c@{}}1,5x$10^{-3}$\\~$\pm$2,1x$10^{-4}$\end{tabular} & \begin{tabular}[c]{@{}c@{}}0.99\\~$\pm$0.01\end{tabular}   & \begin{tabular}[c]{@{}c@{}}0.97\\~$\pm$0.02\end{tabular}                 \\ \hline
\end{tabular}
}
\caption{\label{tab:khi2 table} \textbf{Mean success rate and chi-squared ($\chi^2$) p-value for zero-shot models}, models trained on a single prompt~\PO, models trained on all prompts (\Pall, and models trained with contrastive regularization.} 
\end{table}

\section{Training costs}
We trained the Flan T5 780M and 2.7B, the GPT-Neo 1.3B and the Llama, with eight NVIDIA A100 80GB GPUs, with each LLM instance distributed on one GPU. For the 78M models, we employed four NVIDIA V100 32GB GPUs. Training was conducted with five different seeds in each scenario and environment.  In total, our experiments required 10800 GPU hours on A100 80GB and 6400 GPU hours on V100 32GB.
\end{document}